\documentclass[11pt, fleqn]{article}

\usepackage[utf8]{inputenc}
\usepackage[english]{babel}
\usepackage{hyperref}       % hyperlinks
\usepackage{url}            % simple URL typesetting
\usepackage{amsfonts}       % blackboard math symbols
\usepackage{microtype}      % microtypography
\usepackage{amsmath, amssymb}
\usepackage[a4paper,hscale=0.7,vscale=0.75,centering]{geometry}
\usepackage{fullpage}
\usepackage{authblk}
\usepackage{amsfonts}
\usepackage{graphicx}
\usepackage{float}
\usepackage{xcolor}
\usepackage{subcaption}
\usepackage{wrapfig}
\usepackage{tikz}

\usepackage{algorithm}
\usepackage{fancyvrb}
\fvset{fontsize=\normalsize}
\usepackage[noend]{algpseudocode}
\usetikzlibrary{fit,positioning,bayesnet}
\usetikzlibrary{automata,arrows}
\usetikzlibrary{backgrounds}

\usepackage[acronym,nowarn]{glossaries}
\newacronym{PDDLVM}{PDDLVM}{physics-driven deep latent variable models}
\newacronym{PINN}{PINN}{physics-informed neural networks}
\newacronym{FEM}{FEM}{finite element method}
\newacronym{ELBO}{ELBO}{evidence lower bound}
\newacronym{ADAM}{ADAM}{}

\def\cF{{\mathcal F}}
\def\cB{{\mathcal B}}

\newcommand{\argmax}{\operatornamewithlimits{argmax}}

\def\bkappa{{\boldsymbol \kappa}}

\def\bfSigma{{\mathbf \Sigma}}

\def\A{{\mathbf A}}
\def\Ibf{{\mathbf I}}

\def\y{{\mathbf y}}
\def\z{{\mathbf z}}
\def\kappabf{{\boldsymbol{\kappa}}}

\def\fbf{{\mathbf f}}
\def\ubf{{\mathbf u}}
\def\r{{\mathbf r}}

\def\cB{{\mathcal B}}

\def\cF{{\mathcal F}}

\def\cR{{\mathcal R}}

\def\cG{{\mathcal G}}

\def\bR{{\mathbb R}}

\def\bE{{\mathbb E}}

\def\cB{{\mathcal B}}

\def\NPDF{{\mathcal N}}

\def\cR{{\mathcal R}}
\def\f0{{\mathbf 0}}

\def\md{{\mathrm d}}

\def\b0{{\mathbf 0}}
\def\bpi{{\boldsymbol \pi}}

\definecolor{bred}{rgb}{0.8,0,0}
\definecolor{mustyel}{rgb}{0.88, 0.67, 0.39}
\definecolor{iris}{rgb}{0.35, 0.31, 0.81}
\definecolor{darkchestnut}{rgb}{0.6, 0.41, 0.38}
\newcommand{\cblue}{\textcolor{blue}}

\usepackage{booktabs}

\definecolor{revBcolor}{rgb}{0.76, 0.13, 0.28}
\definecolor{indiagreen}{rgb}{0.07, 0.53, 0.03}
\definecolor{lavenderindigo}{rgb}{0.58, 0.34, 0.92}
\definecolor{hotmagenta}{rgb}{1.0, 0.11, 0.81}
\definecolor{britishracinggreen}{rgb}{0.0, 0.26, 0.15}
\definecolor{caribbeangreen}{rgb}{0.0, 0.8, 0.6}
\definecolor{darkmagenta}{rgb}{0.55, 0.0, 0.55}

\def\bpi{{\boldsymbol \pi}}
\def \bpiFE{{\bpi_{_\text{FE}}}}
\def \bpiCH{{\bpi_{_\text{CH}}}}
\def \bpiNN{{\bpi_{_\text{NN}}}}

\def \ubfFE {{\ubf^{\text{\tiny FE}}}}

\def \fbfFE {{\fbf^{\text{\tiny FE}}}}
\def \fbfWR {{\fbf^{\text{\tiny WR}}}}

\def \bmu {{\boldsymbol \mu}}
\def \bSigma {{\mathbf \Sigma}}
\def \bvarepsilon {\boldsymbol \varepsilon}
\def \g {\mathbf{g}}

\def \bsx {{\boldsymbol x}}
\def \bsu {{\boldsymbol u}}
\title{Fully probabilistic deep models for forward and inverse problems in~parametric~PDEs}

\author[$\cblue{\dagger}$]{Arnaud Vadeboncoeur}
\author[$\cblue{\ddagger}$]{\"Omer Deniz Akyildiz}
\author[$\cblue{\dagger}$]{Ieva Kazlauskaite}
\author[$\cblue{\dagger}$, $\cblue{\star}$]{Mark Girolami}
\author[$\cblue{\dagger}$]{Fehmi Cirak}
\affil[$\cblue{\dagger}$]{Department of Engineering, University of Cambridge}
\affil[$\cblue{\ddagger}$]{Department of Mathematics, Imperial College London}
\affil[$\cblue{\star}$]{The Alan Turing Institute}
\affil[ ]{{\textcolor{blue}{\footnotesize \texttt{av537@cam.ac.uk, deniz.akyildiz@imperial.ac.uk, \{ik394,mag92,fc286\}@cam.ac.uk}}}}

\begin{document}
\maketitle

\begin{abstract}
%% Text of abstract
We introduce a physics-driven deep latent variable model (PDDLVM) to learn simultaneously parameter-to-solution (forward) and solution-to-parameter (inverse) maps of parametric partial differential equations (PDEs). Our formulation leverages conventional PDE discretization techniques, deep neural networks, probabilistic modelling, and variational inference to assemble a fully probabilistic coherent framework. In the posited probabilistic model, both the forward and inverse maps are approximated as Gaussian distributions with a mean and covariance parameterized by deep neural networks. The PDE residual is assumed to be an observed random vector of value zero, hence we model it as a random vector with a zero mean and a user-prescribed covariance. The model is trained by maximizing the probability, that is the evidence or marginal likelihood, of observing a residual of zero by maximizing the evidence lower bound (ELBO). Consequently, the proposed methodology does not require any independent PDE solves and is physics-informed at training time, allowing the real-time solution of PDE forward and inverse problems after training. The proposed framework can be easily extended to seamlessly integrate observed data to solve inverse problems and to build generative models. We demonstrate the efficiency and robustness of our method on finite element discretized parametric PDE problems such as linear and nonlinear Poisson problems, elastic shells with complex 3D geometries, and time-dependent nonlinear and inhomogeneous PDEs using a physics-informed neural network (PINN) discretization. We achieve up to three orders of magnitude speed-up after training compared to traditional finite element method (FEM), while outputting coherent uncertainty estimates.
\end{abstract}

\section{Introduction}
%--------------------------------------------------------------------------------
%
Partial differential equations (PDEs) are central pillars of many fields of science and engineering. They are typically nonlinear and depend on parameters which modulate their behavior, allowing them to describe a wide range of physical and operational conditions. It is thus of great interest to solve PDEs for a range of parameters to obtain numerical simulations of various scenarios. This amounts to solving the PDE for a fixed parameter and is termed \textit{forward problem}, for which an extensive number of solution techniques are available, including the \gls*{FEM}, finite volumes, finite differences, and spectral methods~\cite{ern2004theory, quarteroni2008numerical}. These methods provide a way to move from a parameter to the solution, thus a \texttt{parameter-to-solution} map. The forward problem might be challenging to solve when the PDE is high-dimensional or nonlinear, requiring the numerical methods to have high precision, resulting in a heavy computational burden. On the other hand, given the experimental data of a certain phenomenon, it is often of interest to infer the associated unknown parameters from the experimental data to determine the parameter regimes of the observed system, a setting which is termed the \textit{inverse problem} \cite{tarantola2005inverse}. Recent growing interest in inverse problems is driven by the increased availability of sensor data in science and engineering and advances in machine learning (ML) in exploring high-dimensional spaces. Inverse problems are typically more difficult than forward problems due to under-specification, meaning that small errors in observations may lead to large errors in the model parameter, and the parameters explaining the observations may not be unique. This is particularly true in the presence of observation noise and missing physics~\cite{stuart_2010}, and high-dimensional parameter spaces~\cite{vogel2002computational, arridge_maass_oktem_schonlieb_2019}. 

More specifically, many modern engineering design scenarios require computationally expensive FEM solvers, limiting the amount of experimentation possible within a given time frame/computational budget. Being able to emulate both the forward and the inverse maps allows the practitioners to explore a variety of designs and experimental setups efficiently. For example, given a specific modeling problem (\emph{e.g.} the design of a thin-shell car body) with known governing equations, a designer needs to find the best set of parameters (\emph{e.g.} material properties, geometry, boundary conditions) to obtain a product with desired properties (\emph{e.g.} a structure that does not buckle under a given load). An efficient forward solver allows exploring numerous designs by varying the parameters; this generative process (of proposing design parameters and producing corresponding solutions) would be prohibitively expensive using standard FEM solvers as it requires expensive \textit{Assemble} and \textit{Solve} operations. In addition, observational data may be available for existing designs (either from experiments or high-fidelity simulations) and need to be incorporated into the framework.

%--------------------------------------------------------------------------------
\subsection{Contributions}
%--------------------------------------------------------------------------------
%
Our goal in this paper is to solve simultaneously the forward and inverse problems over a range of parameters, while at the same time building an interpretable, physically meaningful low-dimensional latent space of PDE parameters while providing uncertainty quantification (UQ). Such uncertainty quantification is of great value when modeling parametric PDEs in the forward direction through surrogates because we want to understand the distribution of model confidence between solve instances and within solution domains. UQ in the approximating posterior distributions in inverse problems is also of central importance due to the ill-posedness of many problems ~\cite{stuart_2010}. Most importantly, the Bayesian view in modeling PDEs gives us a principled framework to combine machine learning-based methods with classical numerical schemes in a coherent and reasoned manner and has allowed us to construct a model whereby we can train surrogates to jointly learn a variety of probabilistic maps. To achieve this, we develop a probabilistic formulation called \gls*{PDDLVM} and uniquely blend techniques from numerical solutions of PDEs, deep learning, and probabilistic ML. Notably, the training of the method is free from the costly forward solve operations; in other words, we do not need to generate PDE solutions to act as training data. 

To summarize, we propose a method that (i) simultaneously learns deep probabilistic \texttt{parameter-to-solution} (forward) and \texttt{solution-to-parameter}/\texttt{observation-to-parameter} (inverse) maps, (ii) is free from PDE forward numerical solve operations, (iii) uses the Bayesian view ~\cite{stuart_2010} of regularizing the inverse problem to address their inherent ill-posedness while calibrating model uncertainty in the forward direction, and (iv) allows for the incorporation of various neural networks (NNs) tailored to parametric PDEs (such as \gls*{PINN}s \cite{raissi2019physics}, neural operators \cite{li2020fourier, lu2021learning, anandkumar2020neural, kovachki2023neural}) into a coherent probabilistic framework. We demonstrate this last point by showing that we can embed a \gls*{PINN} network into our variational family; thus our framework also allows for any architecture to be converted into a probabilistic method. 
%Our trained model allows the real-time solution of forward and inverse problems, which is relevant for digital twins, \emph{e.g.}, in engineering design and structural health monitoring.

%--------------------------------------------------------------------------------
\subsection{Related Work}
%--------------------------------------------------------------------------------
%
% \looseness=-1
The challenges associated with forward and inverse parametric PDE problems led to a proliferation of ML-based techniques that aim to aid the forward simulation of physical processes given some parameters and the estimation of parameters that cannot be directly observed (see, \emph{e.g.}, \cite{Han2017solving, Sirignano2018DGM, karniadakis2021physics}). Although the early approaches were primarily based on training ML methods, such as random forests and deep NNs, on observed or simulated physical data~\cite{Raveendran2014Blending} and discovering governing physics laws from the data~\cite{brunton2016discovering}, recent methods focus on ML approaches where physics is an integral part of the model. This is achieved through physics informed loss functions that combine the existing PDE-based parameterizations of physical processes with observed/simulated data~\cite{kim2019deep}, or through differentiable physics~\cite{raissi2019physics, ardizzone2018analyzing, zhao2022learning}. The exceptional success of these methods brought tremendous attention to the emerging field of \textit{physics-informed machine learning} \cite{karniadakis2021physics, ghattas2021learning}. We summarize some of these advances and their relevance to our work below. \\

\noindent\textbf{Physics-informed neural networks (\gls*{PINN}) and related methods.} \gls*{PINN}-based methods~\cite{raissi2018hidden, raissi2019physics, pang2019fpinns, mao2020physics, karniadakis2021physics, cai2022physics} convert the PDE formulation into a \textit{loss function} with either hard or soft constraints \cite{lu2021physics, wang2021understanding}. Most of these methods are based on point-wise evaluation of losses and may struggle to handle complex geometries; see, \emph{e.g.}, \cite{gao2022physics}, for treatment of complex meshes. As loss-based models, \gls*{PINN}s are not inherently probabilistic, unlike the method we develop in the current paper.\\

\noindent\textbf{Learning Forward and Inverse PDE maps.} An alternative approach is to learn mappings between parameters/initial/boundary conditions to solutions, typically requiring parameter-to-solution (\emph{i.e.} input-output) pairs in a supervised setting \cite{lu2021learning, anandkumar2020neural, bhattacharya2020model, wang2020towards, patel2021physics, kropfl2022operator, zhu2018bayesian} which may be unavailable or costly. Further extensions consider the unsupervised case~\cite{wang2021learning, vadeboncoeur2023random}, and couple the model with \gls*{FEM}~\cite{Yin2022Interfacing}.\\

\noindent\textbf{Probabilistic approaches.} Probabilistic formulations of PDE-informed models vary from supervised approaches~\cite{yang2019conditional} and mixed semi-supervised formulations \cite{yang2019adversarial}, to uncertainty quantification procedures based on convolutional neural networks (CNN)~\cite{winovich2019convpde} to Gaussian processes combined with \gls*{FEM}~\cite{cockayne2019bayesian, girolami2021statistical, duffin2021statistical, akyildiz2022statistical}. The most relevant to our work are \cite{kaltenbach2020incorporating, rixner2021probabilistic} where the authors introduce the weighted residual method (which is a generalization of many classical numerical methods for PDEs~\cite{strang1986introduction, chakraverty2019advanced, hatami2017weighted, finlayson2013method, lindgren2009weighted, gerald2004applied}), and treat the residual term as a pseudo-observation \cite{rixner2021probabilistic}. This approach, however, differs from our framework in that we learn the direct mappings between the parameters and the solutions, while \cite{rixner2021probabilistic} introduces an intermediate low-dimensional embedding that is not readily interpretable. Other related approaches consider flow-based probabilistic surrogates~\cite{zhu2019physics}. There are also techniques that use normalizing flows for solving stochastic differential equations in the inverse and forward direction~\cite{guo2022normalizing}. Another line of work integrates physical and deterministic models within variational autoencoders (VAEs). Some of these models focus on the construction of efficient regularized inverse maps \cite{o2019learning, tait2020variational}, while others construct additional latent spaces and relate them to physical observations \cite{qian2021integrating, zhong2023pi}. Related to this line of work, \cite{jacobsen2022disentangling} learn disentangled latent representations of physical processes. Among these, the most relevant to our work is the physics-integrated VAE \cite{takeishi2021physics}, which introduces physics knowledge into a VAE framework. However, our approach is markedly different, as we approximate both forward and inverse maps for parametric PDEs and propose a different variational formulation that constructs a latent space directly characterizing both the solution and the parameters of the physical models.

%--------------------------------------------------------------------------------
\section{Parametric PDEs and their Discretization}\label{sec:technical_background}
%--------------------------------------------------------------------------------
%
We are interested in PDEs defined on a domain $\Omega$ with boundary $\partial \Omega$, generalized as
\begin{subequations}
\begin{align}
    \cG_{z}(u, x) &= f(x), \quad \quad \textnormal{for} \quad \quad x \in \Omega \subset \bR^d, \label{eq:generalPDEA} \\
    \cB_{z}(u, x) &= 0, \quad \, \, \quad \quad \textnormal{for} \quad \quad x \in \partial\Omega, \label{eq:generalPDEB}
\end{align} \label{eq:generalPDE}%
\end{subequations}
where $d \in \{1,2,3\}$, $\mathcal{G}_{z}$ and $\cB_{z}$ are (possibly) nonlinear operators that describe the PDE and its boundary conditions, respectively, $u(x)$ is the solution field, $f(x)$ is the source field, and $z(x)$ is a parameter of the PDE such as the diffusivity field or a scalar like the wave speed. 
%In the case of time-dependent PDEs, we incorporate the time component into the domain variable $x$. 
In the following, we first describe a general way to derive different numerical solution schemes and then introduce \gls*{FEM} and \gls*{PINN}s.

%--------------------------------------------------------------------------------
\subsection{Weighted Residual  Method}\label{sec:technical_background:wrm}
%--------------------------------------------------------------------------------
%
The method of weighted residuals provides a convenient framework for the derivation of many well-known PDE discretization techniques \cite{strang1986introduction, finlayson2013method, lindgren2009weighted, gerald2004applied}. Given a PDE of the form \eqref{eq:generalPDE}, we first define the domain residual 
\begin{align}\label{eq:nonlinearWRM}
    \cR(u,z,f, x) := \cG_{z}(u, x) - f(x).
\end{align}
For the sake of brevity, in the following the boundary term \eqref{eq:generalPDEB} is omitted, \textit{i.e.} it is assumed to be satisfied. We then, multiplying by some arbitrary weight function $w(x)$ and integrating over the domain, obtain a weighted residual functional,
\begin{align}\label{eq:nonlinearWRMIntegrated}
    \int_{\Omega}w(x)\left(\cG_{z}(u, x) - f(x)\right) \md x =  \int_{\Omega}w(x) \cR(u,z,f, x) \md x.
\end{align}
Next we choose a set of test functions
% \rev{$\{w_i(x)\}_{i=0}^m$}
$w_{1:n}(x)$ against which we will be integrating the residual. We further pose that $\hat{u}(x)$ is the \textit{approximant} of the solution field $u(x)$.
% and that $\hat{u}(x)$ is parameterized in terms of the vector $\ubf$.
The form of the relationship between the function $\hat{u}(x)$ that is defined everywhere on $\Omega$ and the vector $\ubf$ describing $\hat{u}(x)$ depends on the approximation technique chosen. Similarly, we pose that the PDE parameter field $z(x)$ and forcing field $f(x)$ are represented by the approximants $\hat{z}(x)$ and $\hat{f}(x)$. 
We then assemble the resulting discrete system of equations into a vectorized form 
\begin{subequations} \label{eq:discreteDiffOp}

\begin{align}
    \A_{\z}(\ubf) &= \left[\int_{\Omega}w_i(x)\cG_{\hat{z}}(\hat{u}, x) \md x\right]_{i=1:n}, \\
    \fbfWR &= \left[\int_{\Omega}w_i(x)\hat{f}(x) \md x\right]_{i=1:n}.\label{eq:fwr}
    % \r   &= \int_{\Omega}w_{1:n}(x)\cR(\hat{u},\hat{z},f, x) \md x.
\end{align}

\end{subequations}
Here, $\A_{\z}: \bR^n \to \bR^n$ is a discretized nonlinear mapping of the differential operator. It allows us to express our discrete residual compactly as
\begin{equation}\label{eq:discreteResidual}
    \r :=\A_\z(\ubf) - \fbfWR,
\end{equation}
with $\r, \ubf, \fbfWR\in \bR^{n}$. This residual description is agnostic to the particular method of discretization used in practice. 
 % then we have PINN-type formulation of the solution field.

%--------------------------------------------------------------------------------
\subsection{Finite Elements and Parameterizations}\label{sec:FEM_Parameterizations}
%--------------------------------------------------------------------------------
%

It is key to leverage the advantages of various discrete representations of solution fields when constructing probabilistic emulators for the solution of physics problems. Finite elements expansions provide flexible function representations which are versatile, but tend to be very high dimensional. By making use of a low dimensional spectral Chebyshev representation of solution fields we can reduce the number of degrees of freedom necessary to describe smooth solutions. Spectral expansions can easily be projected onto arbitrary finite element meshes which is a key characteristic when constructing physics emulators as we typically wish to be mesh independent. We describe the various representations of $u(x)$ as through mappings $\bpi$ as
\begin{align}
    \ubf\underset{\bpiCH}{\mapsto}{\ubfFE}\underset{\bpiFE}{\mapsto}\hat{u}(x),
\end{align}
\begin{figure}[h]
        \centering
        \subcaptionbox{Spectral Chebyshev basis functions $T_i(x)$}{\includegraphics[width=0.45\textwidth]{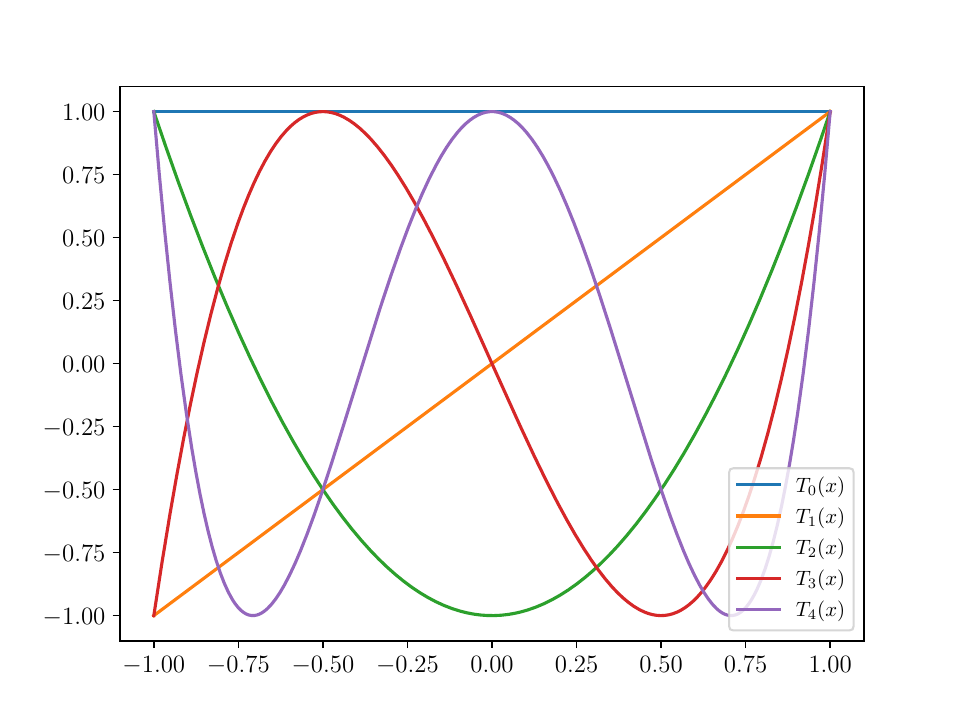}}
        \subcaptionbox{Local \gls*{FEM} linear basis functions $\phi_i(x)$}{\includegraphics[width=0.45\textwidth]{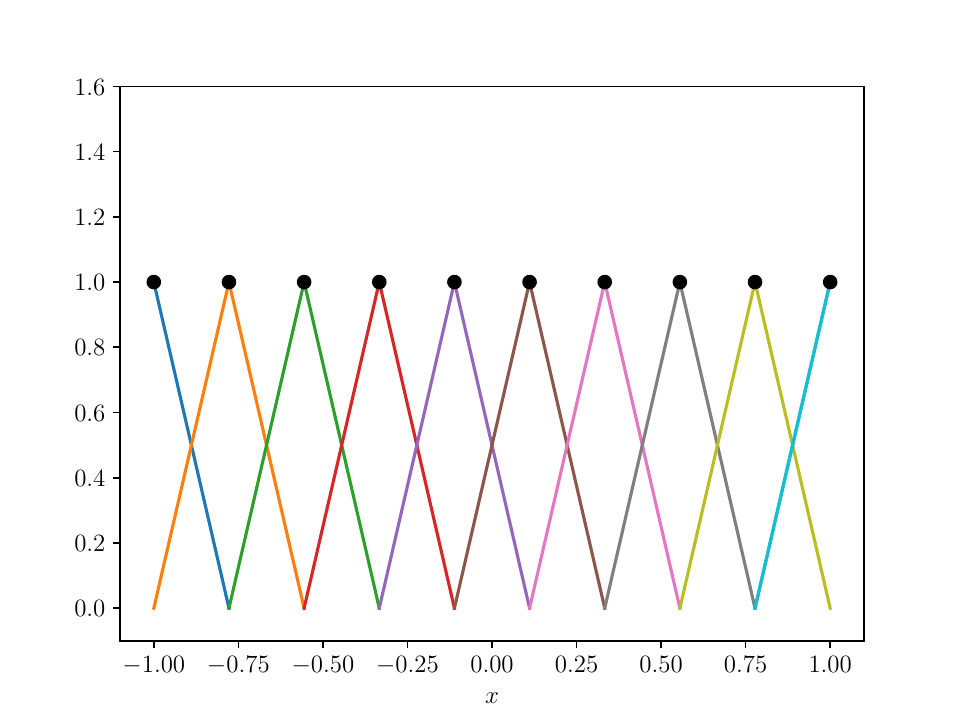}}\\
        \subcaptionbox{Spectral Chebyshev function $\sum_{i}T_i(x)u_i$}{\includegraphics[width=0.45\textwidth]{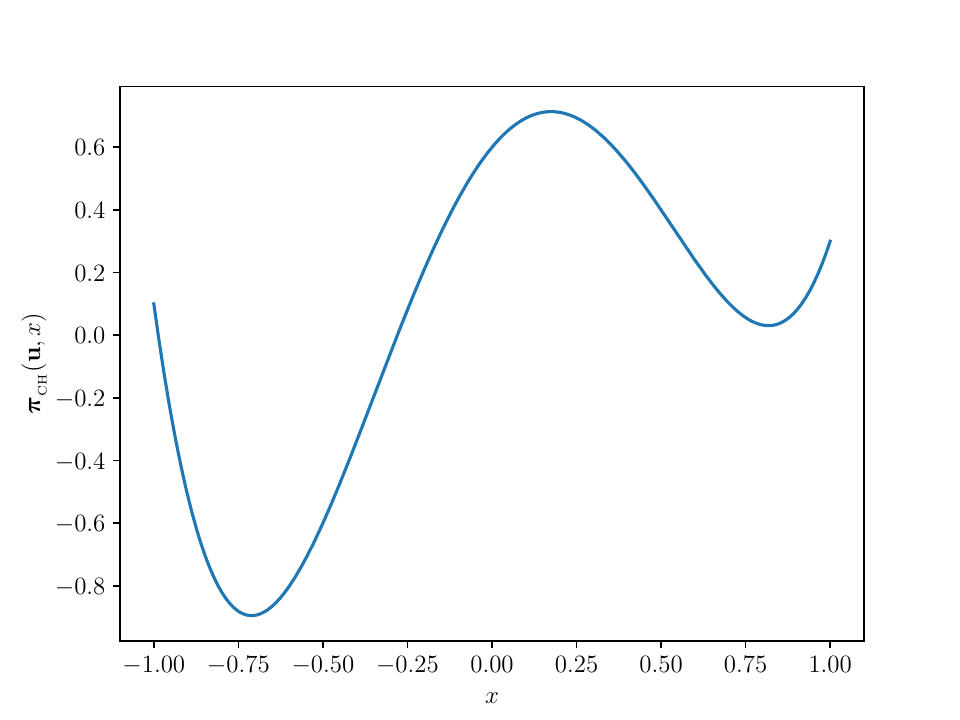}}
        \subcaptionbox{10 node \gls*{FEM} linear interpolation}{\includegraphics[width=0.45\textwidth]{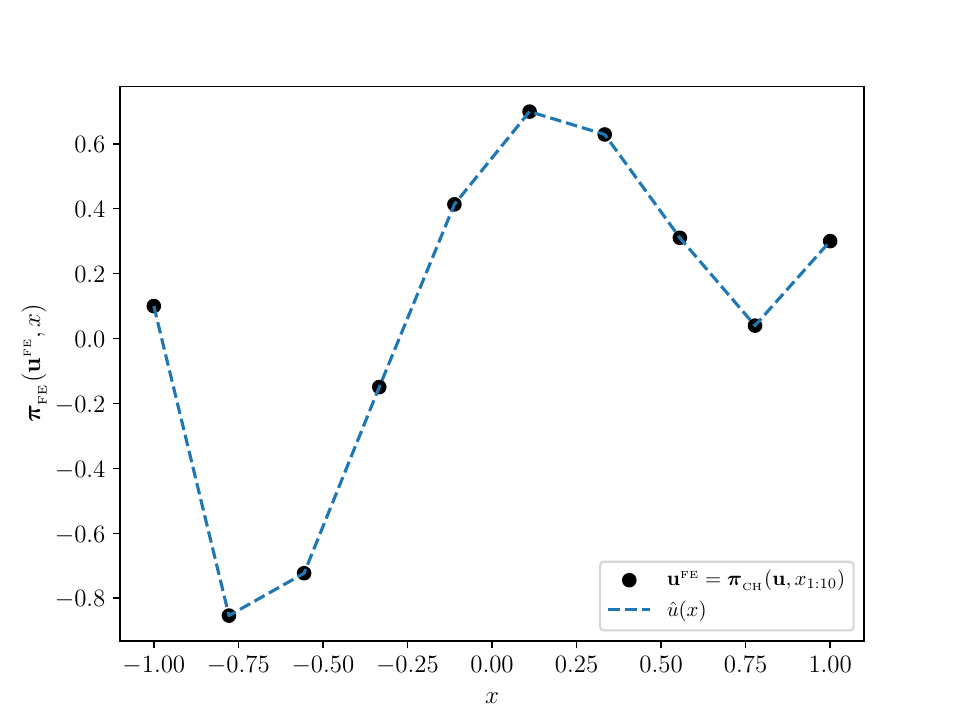}}
    \caption{In (a) we show 5 Chebyshev basis function, in (b) 10 linear \gls*{FEM}  basis functions, in (c) a function composed of 5 Chebyshev basis functions and in (d) its interpolation using 10 linear \gls*{FEM} basis functions. We can see that we require many more terms in a linear interpolation to approximate smooth functions.}
    \label{fig:chebyVfemplot}
\end{figure}
where $\ubf$ denotes the spectral representation of solution fields and $\ubfFE$ are the finite element coefficients. The mapping $\bpiCH$ is a Chebyshev mapping of the spectral coefficients to a set of spatial locations $x_{1:n}$, given by
\begin{align}
\ubfFE &= \bpiCH(\ubf, x_{1:n}) = \left[\sum_i T_i(x_j)u_i\right]_{j=1:n},
\end{align}
where $T_i(x)$ is the $i$th Chebyshev polynomial.
We use a finite element expansion as the approximant solution field
\begin{align}
\hat{u}(x) &= \bpiFE(\ubfFE, x) = \sum_i \phi_i(x){u^{\text{\tiny FE}}_i},
\end{align}
 where $\phi(x)$ the \gls*{FEM} basis functions.
In a Bubnov-Galerkin approach we pose $w_i(x) = \phi_i(x)$. 
The coefficients $u_i$ on the boundaries $\partial \Omega$ are chosen to respect the (Dirichlet) boundary conditions. The mesh allows us to both define the basis functions and describe the geometry of the domain~$\Omega$. We use integration by parts to obtain the \textit{weak form} of \eqref{eq:nonlinearWRMIntegrated}  prior to finite element discretization \cite{brenner2008mathematical}.

It is similarly advantageous to represent our parameter field $z(x)$ through a spectral approximation
\begin{align}
    &\z\underset{\bpiCH}{\mapsto}\hat{z}(x),
\end{align}
which is chosen to be a spectral parameterization as
\begin{align}\label{eq:z_spec_parametrization}
    &\hat{z}(x) = \bpiCH(\z, x) = \sum_i T_i(x)z_i.
\end{align}
We parameterize forcing fields in a similar way to the solution fields through
\begin{align}
    \fbf\underset{\bpiCH}{\mapsto}\fbfFE\underset{\bpiFE}{\mapsto}\hat{f}(x),
\end{align}
each mapping is then given as 
\begin{subequations}
\begin{align}
    &\fbfFE = \bpiCH(\fbf, x_{1:n}) = \left[\sum_i T_i(x_j)f_i\right]_{j=1:n}, \\
    &\hat{f}(x) = \bpiFE(\fbf^{\text{\tiny FE}}, x) = \sum_i  \phi_i(x){f}^{\text{\tiny FE}}_i,
\end{align}
\end{subequations}
where we use \eqref{eq:fwr} to map $\fbfFE$ to $\fbfWR$ through $\hat{f}(x)$ to be used in \eqref{eq:discreteResidual}.
These various parameterizations allow us to represent solution fields, parameter fields, and forcing fields in a \gls*{FEM} mesh invariant manner. We show in Fig~\ref{fig:chebyVfemplot} a representation of 5 spectral Chebyshev basis functions and 9 \gls*{FEM} elements and their local basis functions as well as the interpolation of a Chebyshev function by \gls*{FEM} basis functions.

%--------------------------------------------------------------------------------
\subsection{Physics Informed Neural Networks (PINNs)}
%--------------------------------------------------------------------------------
%
%The use of \gls*{PINN} on the other hand is much simpler. 
To construct a physics informed neural network the trial solution~$\hat u(x)$ is chosen to be the output of a neural network and the spatial locations $x$ are the input. To evaluate the residual~\eqref{eq:nonlinearWRM}, we take the relevant gradients of the trial solution on a grid of (collocation) points and construct the discretized differential equation according to~\eqref{eq:discreteResidual} and~\eqref{eq:discreteDiffOp}. More specifically, we define a mapping $\bpiNN$ between the approximant and the discrete representation of the solution field as
\begin{align}
    \ubf\underset{\bpiNN}{\,\text{\rotatebox[origin=c]{180}{$\mapsto$}}\,}\hat{u}(x_{1:n}).
\end{align}
More precisely, we have $\ubf = \bpiNN(\hat{u}, x_{1:n})$ where
\begin{align*}
    \bpiNN(\hat{u},
x_{1:n}) = [\hat{u}(x_1), \ldots, \hat{u}(x_n)],
\end{align*}
with $x_{1:n}$ is a collection of fixed equidistant collocation points, and 
\begin{align}
     \hat{u}(x) = (\ell_l \circ \hdots \circ \ell_i\circ \hdots \circ \ell_0)(x).
\end{align}
Here, $\ell_i(x) = \boldsymbol{\sigma}(\mathbf{W}_i x+\mathbf{b}_i)$ and $\mathbf{W}$ is a matrix of learnable weights, $\mathbf{b}$ is a vector of learnable biases and $ \boldsymbol{\sigma}$ is a chosen activation function. In this work we only look at PINN implementations with fixed grids. Others works such as \cite{chiu2022can} look at random grid methods, but this is out of the scope of this current work. The framework could be extended to incorporate repeatedly sampled random grids.

In this work we are interested in PINNs for parametric PDEs. For this we pass PDE parameter $\z$ and forcing term $\fbf$ as inputs along with the spatial evaluation coordinate. 
% In this case we represent parameter fields in the same manner as in Sec.~\ref{sec:FEM_Parameterizations} and the forcing field is
To enforce boundary conditions in this framework a second residual is constructed involving the boundary term~\eqref{eq:generalPDEB} and added in a weighted manner to the PDE residual. In our framework, we keep the boundary residual in vectorized form and append it to the domain residual.

%--------------------------------------------------------------------------------
\section{Deep Probabilistic Models for Parametric PDEs}
%--------------------------------------------------------------------------------
%
FEM and PINNs allow to solve the well-posed forward problem for fixed PDE parameters. In settings that require the solution to PDEs for a variety of parameters, the use of either method can be computationally very expensive. On the other hand, the problem of finding $\z$ given some components of $\ubf$ (the inverse problem) may be ill-posed and is generally a harder problem. Solving inverse problems for large collections of observations pertaining to different PDE instances is generally computationally infeasible using classical methods. To overcome these issues, we introduce deep probabilistic  models to directly approximate the maps $(\z, \fbf) \mapsto \ubf$ and $(\ubf, \fbf) \mapsto \z$, while simultaneously providing uncertainty estimates. Once trained, our probabilistic models allow for the online solution of both the forward and inverse problem given a particular forcing $\fbf$. We first describe, in Sec.~\ref{sec:prob_model:dataless}, a method to train a deep probabilistic model to emulate the forward and inverse maps of a PDE without any observed data and without any independent solve operations. 
% (\emph{i.e.} inverting the nonlinear map $\hat \cG_{\z}(u)$).
Building on this, we introduce in Sec.~\ref{sec:prob_model:observed_data} a version of our framework which incorporates observation data.

%--------------------------------------------------------------------------------
\subsection{Probabilistic Model and Variational Approximation without Observed Data}\label{sec:prob_model:dataless}
%--------------------------------------------------------------------------------
%
Following the weighted residual approach we consider the residual $\r \in \bR^n$ in \eqref{eq:discreteResidual} as an auxiliary random variable that is observed and has the value $\r= \mathbf{0}$. 
\begin{figure}
\centering
\begin{center}
\begin{minipage}{.35\textwidth}
\centering
\subcaptionbox{}{
\begin{tikzpicture}
\tikzset{every picture/.append style={background rectangle/.style={fill=black!100, show background rectangle}}}
\tikzstyle{every state}=[fill=black, draw=none, text=white]
\tikzstyle{main}=[circle, minimum size = 6mm, thick, draw =black!100, node distance = 6mm]
\tikzstyle{mainsq}=[rectangle, fill=bred, opacity=.2, text opacity=1, minimum size = 6mm, thick, draw =black!100, node distance = 6mm]
\tikzstyle{connect}=[-latex, thick]
\tikzstyle{box}=[rectangle, draw=black!100]
  \node[main, accepting, double=white] (r_u) {${\r}$};
  \node[main] (z) [above=of r_u] {$\z$};
  \node[mainsq] (beta) [above=of z] {$\beta$-net};
  \node[main] (u) [left=of beta] {$\ubf$};
  \node[main] (f) [right=of beta] {$\fbf$};
  \path (z) edge [connect] (r_u)
  (f) edge (beta)
  (beta) edge [connect] (z)
  (u) edge (beta)
  (u) edge [connect] (r_u)
  (f) edge [connect] (r_u);
\end{tikzpicture}
}
\end{minipage}
\begin{minipage}{.35\textwidth}
\centering
\subcaptionbox{}{
\begin{tikzpicture}
\tikzset{every picture/.append style={background rectangle/.style={fill=black!100, show background rectangle}}}
\tikzstyle{every state}=[fill=black, draw=none, text=white]
\tikzstyle{main}=[circle, minimum size = 6mm, thick, draw =black!100, node distance = 6mm]
\tikzstyle{mainsq}=[rectangle, fill=mustyel, opacity=.2, text opacity=1, minimum size = 6mm, thick, draw =black!100, node distance = 6mm]
\tikzstyle{connect}=[-latex, thick]
\tikzstyle{box}=[rectangle, draw=black!100]
  \node[main] (u) {${\ubf}$};
  \node[mainsq] (alpha) [above=of u] {$\alpha$-net};
  \node[main] (z) [left=of alpha] {$\z$};
  \node[main] (f) [right=of alpha] {$\fbf$};https://www.overleaf.com/project/63230397fdafc321c3ed937d
  \path (z) edge (alpha)
  (f) edge (alpha)
  (alpha) edge [connect] (u);
\end{tikzpicture} }
\end{minipage}
\end{center}
\caption{(a) The probabilistic graphical model with the \textit{inverse} ($\beta$) neural network, where the double rings indicate that $\r$ is an observed vector.  (b) The probabilistic graphical model with the \textit{forward} ($\alpha$) neural network. }
\label{fig:graph}
% \end{wrapfigure}
\label{fig:graphicalModels}
\end{figure}
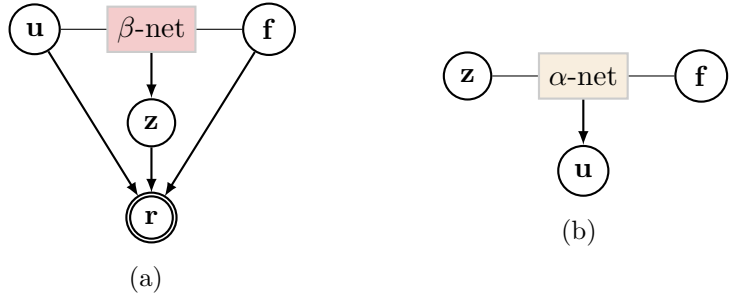
We pose the joint distribution over all variables as
\begin{equation}\label{eq:joint_probability_model_dataless}
    p_{\beta}(\r,\ubf,\z,\fbf) = p(\r | \ubf, \z,  \fbf) p_{\beta}(\z|\ubf,\fbf) p(\ubf) p(\fbf),
\end{equation}
where $\beta$ is the trainable set of parameters in our model, see Figure~\ref{fig:graph}a.  Such a factorization is a modelling choice; the joint distribution could in fact be factorized in different ways so long as the rules of probability are respected. 
The posed factorization was chosen to obtain a conditional distribution $p(\r|\ubf, \z, \fbf)$ of the residual $\r$ as well as a conditional distribution $p_\beta(\z|\ubf, \fbf)$ of $\z$ given the solution $\ubf$ and the forcing $\fbf$. The assumption of this particular model is that $\ubf$ and $\fbf$ are independent when none of the variables in the model are observed. However, it is easy to show that $\ubf$ and $\fbf$  become  conditionally dependent when $\r = \mathbf{0}$ is assumed to be observed~\cite{Bishop2006}. 

We pose the likelihood of the residual as
\begin{equation}\label{eq:likelihood_of_r_u}
    p(\r| \ubf,\z,\fbf) = \NPDF(\r; \A_{\z}(\ubf) - \fbfWR , \mathbf{\Sigma}_\r),
\end{equation}
where $\bfSigma_{\r}$ is the covariance matrix indicating the desired accuracy of the discrete solution $\ubf$ expressed as the deviation of the residual from zero.  
%similar to a statistical \gls*{FEM} (stat\gls*{FEM}) approach \cite{girolami2021statistical, akyildiz2022statistical, duffin2022low}. 
Treating $\r$ as an observed vector and setting $\r = \mathbf{0}$ as in Sec.~\ref{sec:technical_background:wrm}, we aim to maximize its marginal likelihood $p(\r = \mathbf{0})$ by integrating out all other variables.  We next choose a deep probabilistic model of the form $p_\beta(\z | \ubf, \fbf) = \NPDF(\z; \bmu_\beta(\ubf,\fbf), \bSigma_\beta(\ubf,\fbf))$, where $\bmu_\beta$ and $\bSigma_\beta$ are represented by neural networks with the set of parameters $\beta$ consisting of weights and biases. Our model resembles the variational autoencoder (VAE) structure; however, we use a physics-informed model and formulate the parameter as the latent vector~$\z$. The direct correspondence between latent variables and physical parameters enables us to have latent variables with a clear physical interpretation.  

We also use a decoder-like distribution to learn the conditional density $p_\beta(\z | \ubf, \fbf)$ which gives us the flexibility to obtain $\z$ given a solution field $\ubf$ and forcing $\fbf$. To this end, we pose the factorization 
\begin{equation} \label{eq:joint_variational_app_dataless}
    q_\alpha (\ubf, \z, \fbf) = q_\alpha (\ubf | \z, \fbf) p(\z) p (\fbf), 
\end{equation}
with the approximation $q_\alpha(\ubf | \z, \fbf) = \NPDF(\ubf; \bmu_\alpha(\z,\fbf), \bSigma_\alpha(\z,\fbf))$, where $\bmu_\alpha$ and $\bSigma_\alpha$ are represented by neural networks with the set of parameters $\alpha$ consisting of weights and biases, see Fig.~\ref{fig:graphicalModels}b. Inspired by the weighted residual approach introduced in Sec.~\ref{sec:technical_background:wrm} (see also \cite{rixner2021probabilistic}), we maximize the marginal likelihood of the residual by fixing $\r = \mathbf{0}$ and deriving an \gls*{ELBO} that lower bounds the log marginal likelihood $\log p(\r = \mathbf{0})$. 
% In particular, we have
% \begin{align*}
%     \log p(\r = 0) &= \log \int \frac{p_\beta(\r = \mathbf{0}, \ubf, \z, \fbf)}{q_\alpha(\ubf,\z,\fbf)} q_\alpha(\ubf, \z, \fbf) \md \ubf \md \z \md \fbf.
% \end{align*}
 In order to do this, we first note
\begin{align}\label{eq:evidence}
    p(\r = \mathbf{0}) &= \int p_\beta(\r = \mathbf{0}, \ubf, \z, \fbf) \md \ubf \md \z \md \fbf,
\end{align}
where the $p(\r = \mathbf{0})$ acts as \textit{evidence}. Our aim is to maximize the  evidence, i.e. the probability of observing \mbox{$\r = \mathbf{0}$}, by optimising the neural network parameters~$\alpha$ and~$\beta$. To this end, we introduce an instrumental variational approximation and rewrite \eqref{eq:evidence} as
\begin{align}
    p(\r = \mathbf{0}) = \int \frac{p_\beta(\r = \mathbf{0}, \ubf, \z, \fbf)}{q_\alpha(\ubf,\z,\fbf)} q_\alpha(\ubf, \z, \fbf) \md \ubf \md \z \md \fbf,
\end{align}
by dividing and multiplying the integrand in \eqref{eq:evidence} by $q_\alpha(\ubf, \z, \fbf)$. Next, we compute the logarithm of this quantity, since maximizing $\log p(\r = \mathbf{0})$ is equivalent to maximizing $p(\r = \mathbf{0})$ and is more numerically stable. Using Jensen's inequality we can obtain an expression of the form 
\begin{align}
    \log p(\r=\mathbf{0})\geq \cF(\alpha, \beta),
\end{align}
and using \eqref{eq:joint_probability_model_dataless} and \eqref{eq:joint_variational_app_dataless}, we arrive at the lower bound
\begin{align}\label{eq:ELBO_without_data}
        \cF(\alpha,\beta) = \int \log \frac{p(\r = \mathbf{0}|\ubf, \z, \fbf) p_{\beta}(\z | \ubf, \fbf) p(\ubf)}{q_{\alpha}(\ubf | \z, \fbf) p(\z)}q_{\alpha}(\ubf | \z, \fbf) p(\z)p(\fbf) \md \ubf \md \z  \md \fbf.
\end{align}
The \gls*{ELBO} obtained in this manner can also be derived using the Kullback–Leibler divergence between the posterior $p(\ubf, \z, \fbf|\r)$ and the variational approximation $q_\alpha(\ubf, \z, \fbf)$ to it. This link is especially evident in VAE literature \cite{kingma2019introduction}. A full derivation of our \gls*{PDDLVM} framework can be found in \ref{sec:KLDivELBO} where we show that our framework minimizes the KL divergence between the intractable posterior of our latent variable given the pseudo-observed residual $\r = \mathbf{0}$  and a tractable variational approximation in the form $D_{KL}( q_\alpha(\ubf, \z, \fbf)||p_\beta(\ubf,\z,\fbf|\r))$.
Using the derived \gls*{ELBO}, we then seek 
\begin{equation}
    \{\alpha^{\star}, \beta^{\star}\} \in \argmax_{\beta,  \alpha} \cF(\alpha, \beta).
\end{equation}
We note that the $\alpha$ and $\beta$ networks (which parameterize $q_\alpha(\ubf | \z, \fbf)$ and $p_\beta(\z | \ubf, \fbf)$, respectively) have intuitive functionalities (see Fig.~\ref{fig:graph} for the graphical models). In particular, the $\alpha$ network provides a \texttt{parameter-to-solution} (forward) map, while the $\beta$ network provides a \texttt{solution-to-parameter} (inverse) map, both with associated uncertainty estimates.
Modern variational inference techniques rely on maximizing the lower bound $\cF(\alpha, \beta)$, instead of the true evidence $\log p(\r = \mathbf{0})$ to learn the parameters $(\alpha, \beta)$. To compute \eqref{eq:ELBO_without_data} we express the integral as an expectation and use Monte-Carlo approximation of this expectation. The $N$-sample Monte Carlo \gls*{ELBO} estimate of the gradient of $\cF(\alpha, \beta)$ can be obtained by sampling from $p(\z)$, $p(\fbf)$ and $q_\alpha(\ubf | \z, \fbf)$ (using the reparameterization trick~\cite{kingma2013auto}). For this, we rewrite \gls*{ELBO} \eqref{eq:ELBO_without_data} as a generic integral
\begin{align}
    \cF(\alpha, \beta) = \bE_{q_\alpha(\ubf | \z, \fbf) p(\z) p(\fbf)}[\mathsf{v}_{\alpha,\beta}(\ubf, \z, \fbf)],
\end{align}
where
\begin{align}
    \mathsf{v}_{\alpha,\beta}(\ubf, \z, \fbf) = \log \frac{p(\r = \mathbf{0}|\ubf, \z, \fbf) p_{\beta}(\z | \ubf, \fbf) p(\ubf)}{q_{\alpha}(\ubf | \z, \fbf) p(\z)}.
\end{align}
Given that $\bvarepsilon \sim \NPDF(\mathbf{0}, \mathbf{I})$, we can sample $\ubf \sim q_\alpha(\ubf | \z, \fbf)$ by sampling $\bvarepsilon$ and computing
\begin{align}\label{eq:reparam_trick_dataless}
    \mathsf{w}_\alpha(\z, \fbf, \bvarepsilon) = \bmu_\alpha(\z, \fbf) + \bSigma^{1/2}_\alpha(\z, \fbf) \bvarepsilon,
\end{align}
and write
\begin{align}
        \cF(\alpha, \beta) &= \bE_{q_\alpha(\ubf | \z, \fbf) p(\z) p(\fbf)}[\mathsf{v}_{\alpha,\beta}(\ubf, \z, \fbf)]
        = \bE_{q(\bvarepsilon) p(\z) p(\fbf)}\left[\mathsf{v}_{\alpha,\beta}(\mathsf{w}_\alpha(\z, \fbf, \bvarepsilon), \z, \fbf)\right]\nonumber.
\end{align}
To construct the Monte Carlo \gls*{ELBO}, we now sample $\bvarepsilon^{(j)} \sim q(\bvarepsilon) , \z^{(j)} \sim  p(\z), \fbf^{(j)} \sim p(\fbf)$ for $j = 1,\ldots, N$ and obtain
\begin{align}\label{eq:MonteCarloElbo_dataless}
    \cF^N(\alpha, \beta) = \frac{1}{N}\sum_{j=1}^N \mathsf{v}_{\alpha,\beta}(\mathsf{w}_\alpha(\z^{(j)},\fbf^{(j)},\bvarepsilon^{(j)}), \z^{(j)}, \fbf^{(j)}).
\end{align}
The rest of the training is done by computing the gradient of this stochastic loss w.r.t. $\alpha,\beta$ and running a variant of gradient descent, such as Adam \cite{kingma2015adam}. In many practical applications (and in this paper), $N = 1$, \emph{i.e.}, a one-sample estimate of the gradient is utilized \cite{kingma2019introduction}. We note that one of the strengths of this method is that both linear and nonlinear PDEs can be treated the same way as we only need to evaluate the residual term and its gradient. 

\begin{algorithm}[h]
  \caption{Pseudocode for \gls*{PDDLVM}\label{alg:training}}
  \begin{algorithmic}
    \State Initialize: $\alpha_0$, $\beta_0$, $T$ (number of iterations), $N$ (number of Monte Carlo samples).
    \For{$t=1, \ldots,  T$}   \Comment{Gradient descent.}
    \For {$i = 1, \ldots, N$} \Comment{Monte Carlo estimate of the loss \eqref{eq:ELBO_without_data}.}
      \State Sample $\z^{(i)} \sim p(\z)$
      \State Sample $\fbf^{(i)} \sim p(\fbf)$
      \State Sample $\ubf^{(i)} \sim q_{\alpha_{t-1}}(\ubf | \z^{(i)}, \fbf^{(i)})$
     \EndFor
      \State Compute $\cF^N(\alpha, \beta)$ using the samples.  
      \State $(\alpha_t, \beta_t) \gets \textnormal{OPTIMISER}(\alpha_{t-1}, \beta_{t-1}, \cF^N(\alpha, \beta))$ \Comment{Update parameters.}
    \EndFor
  \end{algorithmic}
\end{algorithm}

%--------------------------------------------------------------------------------
\subsection{Probabilistic Model and Variational Approximation with Observed Data}\label{sec:prob_model:observed_data}
%--------------------------------------------------------------------------------
%
We now introduce observations into our model to cover data-centric scenarios, as well as to handle multiple forcing terms. In this section, we assume that the $\beta$-network yielding the distribution $p_\beta (\z| \ubf, \fbf) $ is pretrained as described in Sec.~\ref{sec:prob_model:dataless}. We denote the learned $\beta$ network parameter by setting $\beta = \beta_\star$. For a given observation set $\{\y_i \}_{i=1}^{m}$, where $\y_i\in \bR^{n_y}$, and a corresponding solution set $\{ \ubf_i \}_{i=1}^{m}$, we introduce the observation likelihood
\begin{equation}
    p(\y_i | \ubf_i) = \NPDF(\y_i; \g(\ubf_i), \bfSigma_\y), \quad\quad \text{for } \quad i = 1,\ldots,m,
\end{equation}
where $\g: \bR^n \to \bR^{n_y}$ is a \textit{known} observation operator and $\bSigma_\y$ is the noise covariance.
We pose our probabilistic model for a tuple $(\y_i, \ubf_i)$ as
\begin{equation}
    p(\y_{i}, \ubf_i) = p(\y_i|\ubf_i)p(\ubf_i).
\end{equation}
Note in this case that our probability model is fully specified, with no trainable parameters.  We next describe our variational approximation as
\begin{equation}
    q_\phi(\ubf_i | \y_i) = \NPDF(\ubf_i; \bmu_\phi(\y_i), \bSigma_\phi(\y_i)).
\end{equation}
% Our variational family in this case introduces a $\phi$-network which aims at learning the \texttt{observation-to-solution} map. 
The marginal likelihood of the data is given by
\begin{align}\label{eq:evidence_data}
    p(\y_i) &= \int p(\y_i, \ubf_i) \md \ubf_i.
\end{align}
Incorporating the variational distribution $q_\phi(\ubf_i|\y_i)$ for the \texttt{observation-to-solution} map we can write for the log marginal likelihood of the data
\begin{align}
    \log p(\y_i) &= \log \int \frac{p(\y_i, \ubf_i)}{q_\phi(\ubf_i|\y_i)} q_\phi(\ubf_i|\y_i) \md \ubf_i.
\end{align}
Using Jensen's inequality this expression can be bounded by the \gls*{ELBO}
\begin{equation}\label{eq:ELBO_with_data}
   \cF_i(\phi) = \int \log \frac{p(\y_i|\ubf_i) p(\ubf_i)}{q_{\phi}(\ubf_i|\y_i)} q_{\phi}(\ubf_i | \y_i)  \md \ubf_i .
\end{equation}
Given the dataset $\{\y_i\}_{i=1}^m$, the full \gls*{ELBO} is $\cF(\phi) = \sum_{i=1}^m \cF_i(\phi)$.
Once trained, the variational distribution~$q_\phi (\ubf_i | \y_i)$ is sufficient to obtain an encoding model for observations. The lower bound can be computed in a similar manner to the \gls*{ELBO} in Sec.~\ref{sec:prob_model:dataless} with
\begin{align}
    \cF_i(\phi) = \bE_{q_\phi(\ubf_i | \y_i)}[\mathsf{v}_{\phi}(\y_i,\ubf_i)],
\end{align}
where
\begin{align}
    \mathsf{v}_{\phi}(\y_i,\ubf_i) = \log \frac{p(\y_i|\ubf_i) p(\ubf_i)}{q_{\phi}(\ubf_i|\y_i)} q_{\phi}(\ubf_i | \y_i),
\end{align}
which can also be computed through the reparametrization trick detailed in \eqref{eq:reparam_trick_dataless} as 
\begin{align}\label{eq:reparam_trick_data}
    \mathsf{w}_\phi(\y_i, \bvarepsilon) = \bmu_\phi(\y_i) + \bSigma^{1/2}_\phi(\y_i) \bvarepsilon,
\end{align}
and write
\begin{align}
        \cF_i(\phi) &= \bE_{q_\phi(\ubf_i | \y_i)}[\mathsf{v}_{\phi}(\y_i,\ubf_i)] =  
        \bE_{q(\bvarepsilon)}\left[\mathsf{v}_{\phi}(\y_i, \mathsf{w}_\phi(\y_i, \bvarepsilon))\right]\nonumber.
\end{align}
This can be approximated with Monte-Carlo sampling as done in \eqref{eq:MonteCarloElbo_dataless}.

To realize the \texttt{observation-to-parameter} map, we can then use the already learned $\beta_\star$-network to marginalize over $\ubf$ as 
\begin{equation}\label{eq:predictiveZoverU}
    p(\z|\y, \fbf) = \int q_{\phi_\star}(\ubf|\y)p_{\beta_\star}(\z|\ubf, \fbf)\md \ubf.
\end{equation}
The marginalization of $\z$ over $\ubf$ can also be easily extended to include the forcing $\fbf$. We stress that the method we propose only uses data when we are inverting some observable map from a dataset, this extension does not learn the physics of a problem but rather some transformation that is over the underlying physics. The physics are learned using the algorithm described in Sec.~\ref{sec:prob_model:dataless} and this probabilistic physics model can then be used for combined inference as shown in \eqref{eq:predictiveZoverU}.

%--------------------------------------------------------------------------------
\subsection{Algorithm and Implementation Details}
%--------------------------------------------------------------------------------
% In this section, we summarize some key algorithmic aspects of our framework.
%
\noindent\textbf{Priors and Monte Carlo \gls*{ELBO}.} In all formulations, we assume a flat prior $p(\ubf) \propto 1$ for the solution field. We choose $p(\fbf)$ and $p(\z)$ as uniform or normal distributions (see Sec.~\ref{sec:experiments}). We estimate the \gls*{ELBO}s in \eqref{eq:ELBO_without_data} using sampling from $\z \sim p(\z)$ and $\fbf \sim p(\fbf)$, $\ubf \sim q_\alpha(\ubf | \z, \fbf)$ (using the reparameterization trick \cite{kingma2013auto})) and optimize with Adam \cite{kingma2015adam}. When multiple observations exist, \emph{i.e.} if $m \gg 1$, we use only one-sample estimate of \gls*{ELBO}, resulting in a doubly-stochastic gradient. See Algorithm~\ref{alg:training} for the full algorithm. We also provide details in Sec.~\ref{sec:experiments} where necessary.
\begin{figure}[h]
% \begin{figure}
    \centering
    \subcaptionbox{Architecture for \gls*{FEM}-\gls*{PDDLVM}}{\includegraphics[angle =0, width=\textwidth]{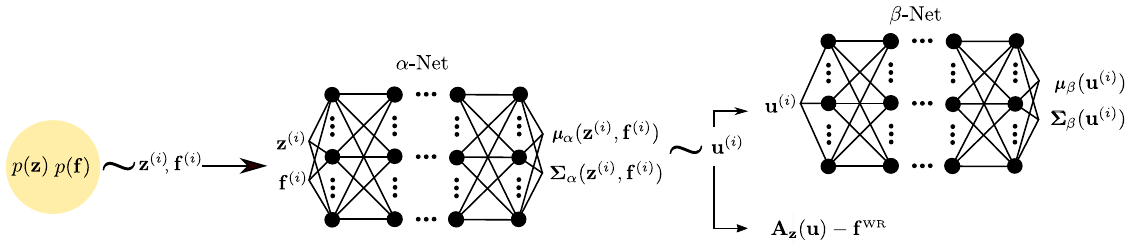}}
    \subcaptionbox{Architecture for \gls*{PINN}-\gls*{PDDLVM}}{\includegraphics[angle =0, width=\textwidth]{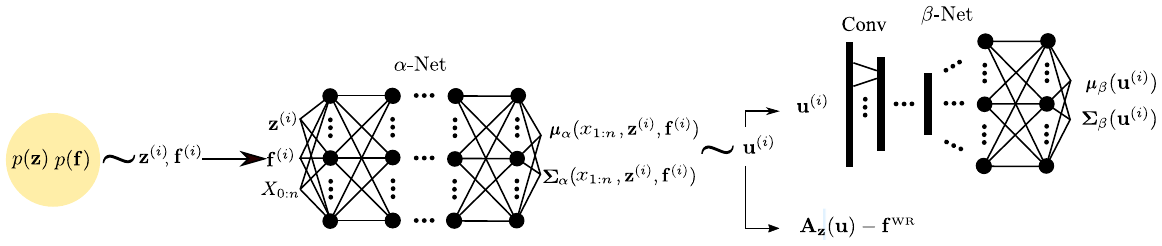}}
    \caption{The architecture of both the \gls*{FEM}-\gls*{PDDLVM} and the \gls*{PINN}-\gls*{PDDLVM}. The large ``$\sim$'' symbol denotes the drawing of random samples from the distribution (or from the distribution described by the given parameters) on the left of it. The two variants of \gls*{PDDLVM} differ in the fact that for the \gls*{FEM} version we reparametrize the samples $\ubf^{(i)}$ with a global expansion which in the shown examples are Chebyshev expansions of various orders. For the \gls*{PINN}-\gls*{PDDLVM} version we instead evaluate the network at the collocation gridpoints and sample directly in the solution space of the domain, this yields the simplification $u^{(i)}_j = \hat{u}^{(i)}(x_j)$. In the \gls*{PINN} version the $\beta$-Net is implemented as a convolutional neural network taking in the trial solution evaluated on the grid, it can in essence be interpreted as an image.}\label{fig:archPDDLVM}
\end{figure}
In Fig.~\ref{fig:archPDDLVM} we show diagrams depicting the flow of samples in the algorithm. With all the quantities computed through the given samples we can then evaluate all the relevant terms in the $\gls*{ELBO}$.

%--------------------------------------------------------------------------------
\section{Examples}\label{sec:experiments}
%--------------------------------------------------------------------------------
%
We demonstrate our methodology on six selected examples\footnote{The code is made available at  \href{https://github.com/ArnaudVadeboncoeur/PDDLVM}{https://github.com/ArnaudVadeboncoeur/PDDLVM}}. In all experiments, we choose a diagonal covariance for the networks, although other choices can be explored \cite{kingma2019introduction}. Furthermore, we make use of a bounded re-parameterization of the log variance of the neural networks \cite{dehaene2021re}. We note that our approach is extrusive in the sense that it needs only the solution and its gradient but is agnostic to the inner workings of the \gls*{FEM} package.  The examples in Sec.~\ref{sec:LinearPoisson}, Sec.~\ref{sec:OneDPoissonNonlinear}, and Sec.~\ref{sec:3DshellObject} all use a \gls*{FEM}-\gls*{PDDLVM} residual formulation and the output of the neural network is parameterized with a Chebyshev expansion. The further sections are based on the \gls*{PINN} formulation of the residual. A summary of relevant information for each experiment can be found in~\ref{app:info_tables}. 

%--------------------------------------------------------------------------------
\subsection{1D Linear Poisson}\label{sec:LinearPoisson}
%--------------------------------------------------------------------------------
%
We use the formulation in Sec.~\ref{sec:prob_model:dataless} to train the $\alpha$ and $\beta$ networks on a linear Poisson equation with Dirichlet boundary conditions 
\begin{align}
   -\nabla\cdot(\kappa(x)\nabla u(x)) &= f(x), \quad x\in (-1,1),
   \label{eq:LinearPoisson} \\
   u(-1) = a, \quad & u(1) = b. \nonumber
\end{align}
we learn the \texttt{parameter-to-solution} map jointly with the \texttt{solution-to-parameter} map with uncertainties.

When using~\eqref{eq:z_spec_parametrization} to represent a physical quantity which is strictly positive we compose the spectral mapping $\bpiCH$ with a Softplus transformation of the form $\log(1+\exp(x))$. As the transformation is nonlinear, we make use of the unscented transform \cite{wan2000unscented} for estimation of the first and second moments of the probability density when mapping the probability of $\z$ onto the domain $\Omega$. When we project $\z$ in Chebyshev coefficients onto the finite element mesh as $\hat{z}(x_{1:n})$, we can transform its probability distributions as $p(\hat{z}(x_{1:n})) = \NPDF(\hat{z}(x_{1:n}); \bmu_\beta \mathbf{T}^\top, \mathbf{T} \mathbf{\Sigma}_\beta \mathbf{T}^\top)$, where $\mathbf{T}$ is the Chebyshev Vandermond matrix at the mesh locations $x_{1:n}$. The choice of Chebyshev polynomials for this expansion is motivated by their stability (as opposed to monomial or Lagrange polynomials, which suffer from the Runge's phenomenon) and the ease with which they approximate constant, linear, and quadratic functions. Other expansions such as Fourier series might require many terms to approximate these simple functions with accuracy.

%--------------------------------------------------------------------------------
\subsubsection{Learning Without Observations}\label{sec:joint_learning_without_obs}
\begin{figure}[h]
        \centering
        \subcaptionbox{FE Solver}{\includegraphics[width=0.45\textwidth]{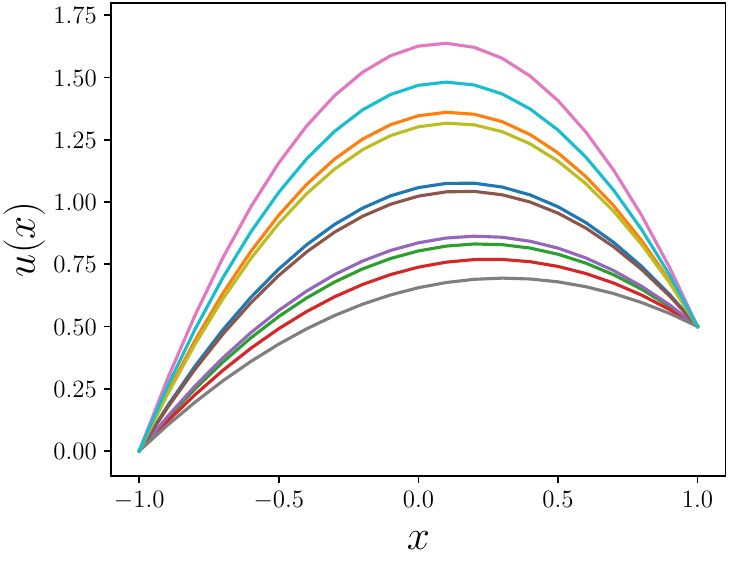}}
        \hspace{3em}
        \subcaptionbox{\gls*{PDDLVM} Forward }{\includegraphics[width=0.45\textwidth]{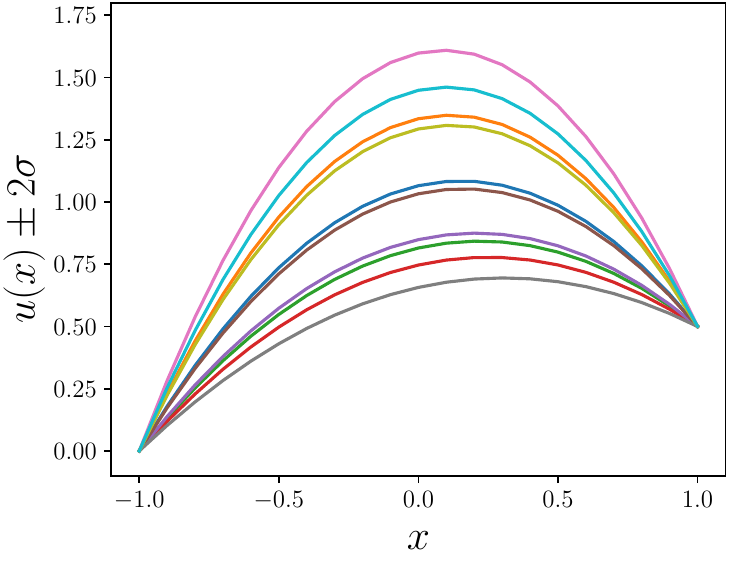}}\\
        \subcaptionbox{True $\kappa(x)$}{\includegraphics[width=0.45\textwidth]{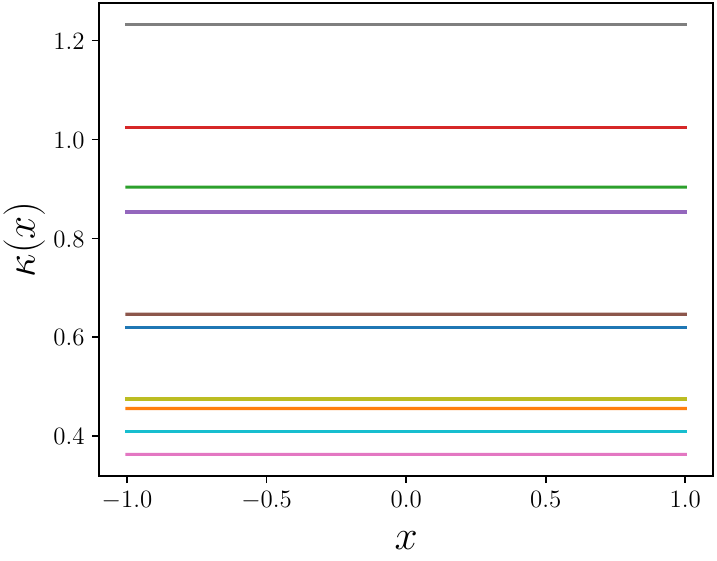}}
        \hspace{3em}
        \subcaptionbox{\gls*{PDDLVM} Inverse}{\includegraphics[width=0.45\textwidth]{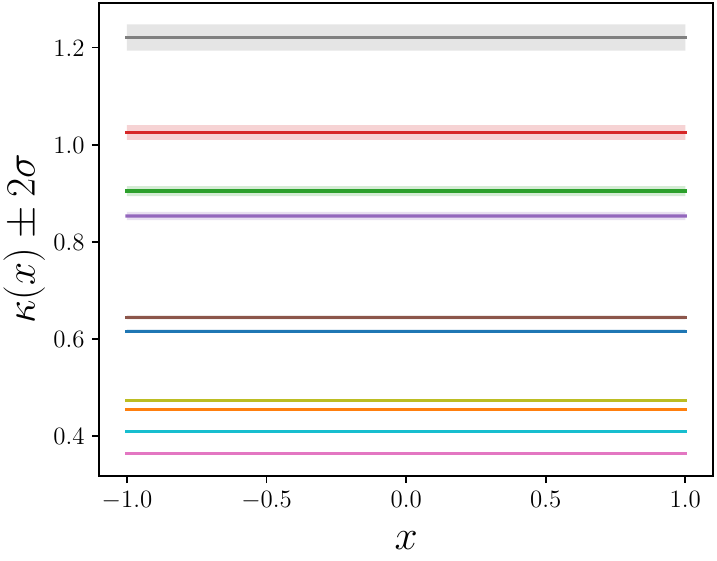}}
    \caption{Results for the 1D linear Poisson equation. (a) The true solutions given by a linear \gls*{FEM} solver for 10 different constant functions of $\kappa(x)$. (b) The predicted solution with $2\sigma$ interval given by the \gls*{PDDLVM} for those same functions $\kappa(x)$. (c) The 10 true $\kappa(x)$ functions used to compute the solution fields in plots (a) and (b). (d) The  \gls*{PDDLVM} reconstruction of $\kappa(x)$ from the solution field of plot (a) (identical functions means a perfect reconstruction of the $\kappa(x)$ field). }
    \label{fig:LinearPoisson}
\end{figure}

%--------------------------------------------------------------------------------
%
In this section we demonstrate the use of our method on the linear Poisson problem without observations as in Sec.~\ref{sec:prob_model:dataless}. To measure the accuracy of our method after training, we compute the \gls*{FEM} ground truth solution, however, these are \textit{not} used during training. The residual is marginalized over the prior of $\z$ and $\fbf$ which allow us to learn the PDE over the prior ranges. In this example, our interpretable latent variable $\z = \kappa$ denotes the coefficients of the $\kappa(x)$ expansion. To ensure positivity of the diffusion field, we pass $\kappa(x)$ through a Softplus function of the form $\log(1+\exp(\kappa(x)))$. In Fig.~\ref{fig:LinearPoisson} the values for the boundary conditions $a$ and $b$ are kept fixed at 0 and 0.5. We plot the results from the \gls*{PDDLVM} inference for 10 random samples drawn from the prior of ${\kappa}$. The diffusivity field $\kappa(x)$ and forcing $f(x)$ is parameterized as a constant function and the solution field is given by a $4^\text{th}$ order Chebyshev expansion. The prior over $\kappa$ is $\NPDF(\kappa; 0, 1)$. We use a NN architecture of 50 nodes for 3 hidden layers. We use a ``swish'' activation function and train each model for $2\times 10^5$ iterations with Adam and a learning rate of $10^{-3}$ decayed by half 10 times over the training time. The joint training of the $\alpha$ and $\beta$ networks takes 218 seconds.
In Fig.~\ref{fig:LinearPoisson}b the shaded region for 2 standard deviation is very small as the forward problem is well determined and so the confidence region is highly concentrated around the mean function. In contrast to this, the inverse problem is more ill-posed and this is reflected by the larger uncertainty bands around the mean of the diffusion field in Fig.~\ref{fig:LinearPoisson}.

\begin{figure}
    \centering
    \begin{subfigure}{.45\textwidth}
      \centering
      \includegraphics[width=1\linewidth]{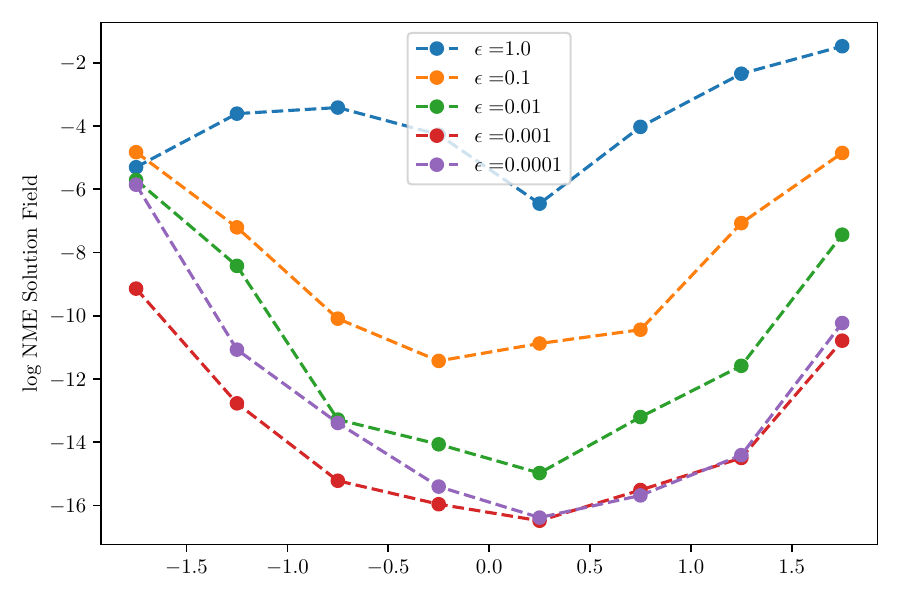}
      \caption{$u(x)$}
      \label{fig:sfigU}
    \end{subfigure}%
    \hspace{3em}
    \begin{subfigure}{.45\textwidth}
      \centering
      \includegraphics[width=1\linewidth]{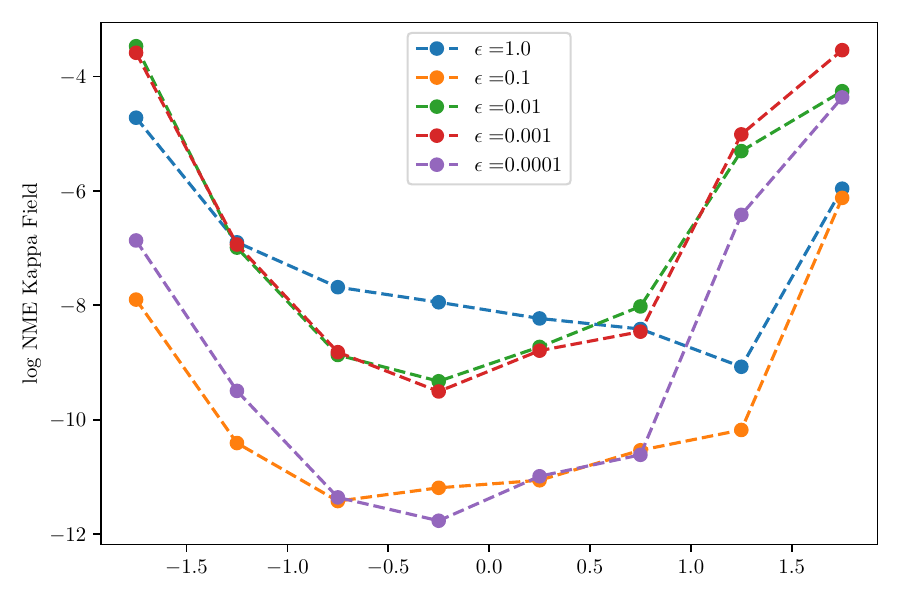}
      \caption{$\kappa(x)$}
      \label{fig:sfigK}
    \end{subfigure} 
    \begin{subfigure}{.45\textwidth}
      \centering
      \includegraphics[width=1\linewidth]{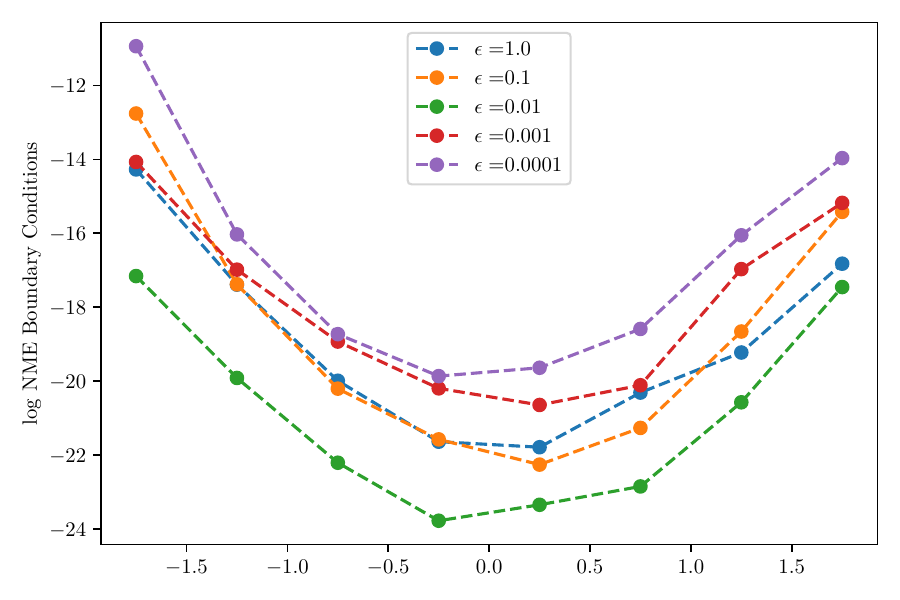}
      \caption{Boundary conditions}
      \label{fig:sfigB}
    \end{subfigure}
\caption{The MNSE of PDDLVM predictions when compared to the ground truth. The algorithm was trained with a prior of $p(\kappa)=\NPDF(0, 0.5)$ and was tested with 100 random draws of new $\kappa$ from a narrow uniform testing distribution of width 0.5. We show the variation of error as we slide this testing distribution from values of -2 to 2. The points on the graph denote the center of this narrow testing distribution, the testing distribution extends in both direction by 0.25. In this manner, we compare the model's predictions to the \gls*{FEM} solution as well as the true values of $\kappa$ and the true values of the boundary conditions. We can see that as we go farther from regions of high prior density of $\kappa$, the error increases and vise-versa.
We plot the $\log$ MNSE for (a) the inferred solution field (forward problem), (b) the reconstructed diffusion field from the solution (inverse problem), and (c) the boundary conditions. This is computed for a variety of parameters $\epsilon_\r$. The boundary conditions are kept fixed throughout this experiment via a Dirac prior during training. However, we treat the value of the boundary conditions as variable to be inferred. We can see that error in boundary conditions inferred from inverting the solution field still varies with respect to how close we sample $\kappa$ to the main probability mass of $p(\kappa$).}
\label{fig:PoissonLinearLOG-MNSE}
\end{figure}

Using this model, we study the effect of varying the $\epsilon_\r$ value of the residual covariance, i.e. $\boldsymbol \Sigma_\r = \epsilon_\r^2 \boldsymbol I$, controlling the solution accuracy. In Fig.~\ref{fig:PoissonLinearLOG-MNSE} we plot the $\log$ Mean Normalized Squared Error (MNSE) \eqref{eq:MNSE} for 100 samples of solution fields, diffusion field, and boundary conditions for a variety of ${\kappa}$ from the prior. Similarly, in the Appendix Fig.~\ref{fig:PoissonLinearMNSEDistributionK} we plot the mean normalized squared error for 100 independent draws of $\kappa$ from various ranges of values. The error for the proposed solution field is computed with respect to a true \gls*{FEM} solution. For the reconstructed quantities $\kappa$ and the boundary conditions, we compare the generated samples with the reconstructed one. As expected, the MNSE tends to increase as we move away from the regions of highest probability density of our prior distribution of $p(\kappa)$. We see that irrespective of the value of $\epsilon_\r$, the error increases for the solution field coefficients, the diffusivity field coefficients, and the boundary conditions as we move away from the main probability mass of the prior. As can also be seen in Fig.~\ref{fig:Poisson1DELBOforEpsi}, changes in the value for $\epsilon_\r$ tempers the optimization surface which will affect the training of the algorithm. We can conclude from this that there is no single correct setting of $\epsilon_\r$ but in effect is a practitioner's choice which will impact the learning process.

%--------------------------------------------------------------------------------
\subsubsection{Observable Map Inversion}\label{sec:obs_map_inversion}
%--------------------------------------------------------------------------------
%
\begin{figure}
        \centering
        \subcaptionbox{$\y$}{\includegraphics[width=0.45\textwidth]{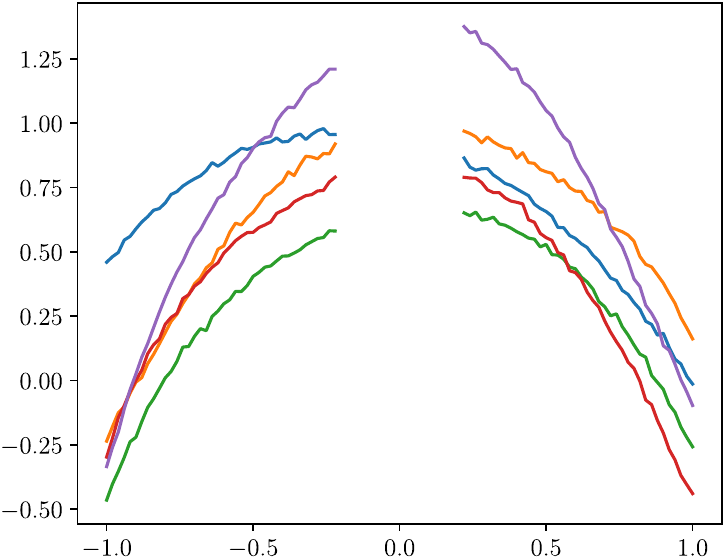}}
        \hspace{3em}
        \subcaptionbox{True $u(x)$}{\includegraphics[width=0.45\textwidth]{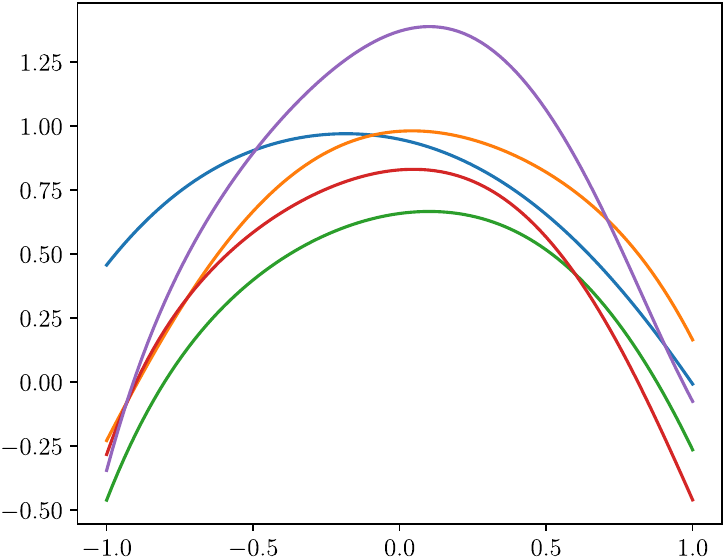}}
        \\
        % \subcaptionbox{True $\kappa(x)$}{\includegraphics[width=0.43\textwidth]{figs/OneDPoissonInvertObservableMap/TrueKappa_rev_resize.pdf}}
        \subcaptionbox{True $\kappa(x)$}{\includegraphics[width=0.45\textwidth]{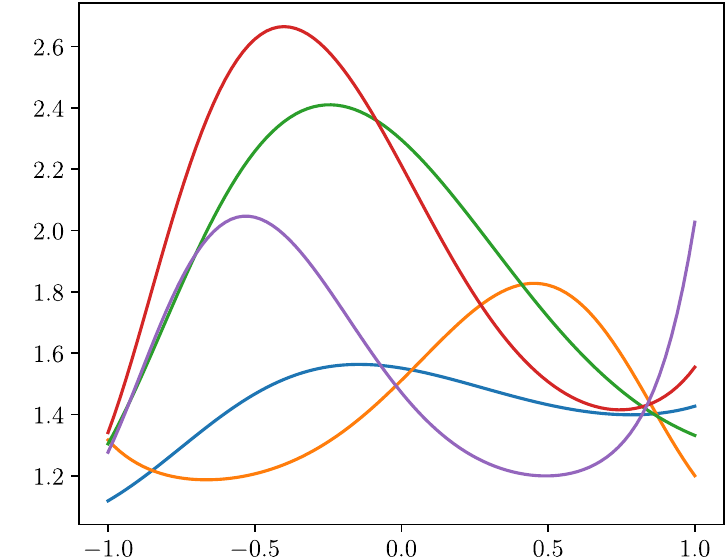}}
        \hspace{3em}
        \subcaptionbox{$q_{\phi}(\ubf|\y)$}{\includegraphics[width=0.45\textwidth]{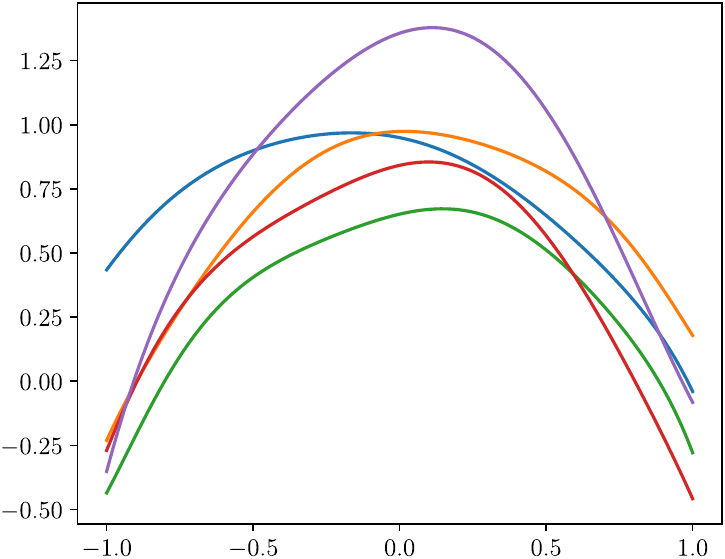}}
        \\
        \subcaptionbox{$p_{\beta^\star}(\z|\bar{\ubf}, \fbf)$}{\includegraphics[width=0.45\textwidth]{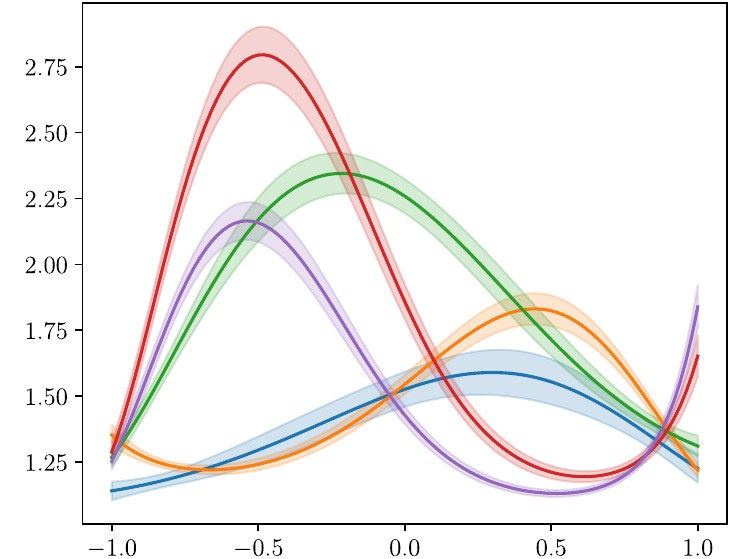}}
    \caption{Example of missing data. (a) Five data samples not observed during training. (b) The true $u(x)$ used to generate the synthetic data in (a). (c) The true parameter coefficient used to generate (b). (d) The predictive distribution of $\ubf$ given an observation $\y$. (e) The inferred distribution over $\kappa(x)$ given the inferred mean of $\ubf$ in (d) and the forcing. The variable $\bar{\ubf}$ denotes the mean of the predicted distribution given by $q_\phi(\ubf|\y)$. }
    \label{fig:ObsMapExamples}
\end{figure} 
We make use of Sec.~\ref{sec:prob_model:observed_data} to learn to invert a noisy mapping from the measurements to the solutions in the presence of observations. Here the governing equation is \eqref{eq:LinearPoisson} with varying boundary conditions and the observable forward map is a truncation operation of the middle 20 values on the observable grid of 100 mesh points. We use a set of 100 observations with a noise covariance of $\mathbf{\Sigma_y} = \sigma_y ^ 2 \mathbf{I}$, where $\sigma_y = 0.01$. The $\alpha_{\star}$ and $\beta_\star$ networks are pretrained on the linear Poisson equation with $\kappa(x)$ expressed as a $3^{\text{rd}}$ order Chebyshev expansion, variable boundary conditions, and variable constant function forcing. The model $q_{\phi}$ reconstructs $\ubf$ as Chebyshev coefficients from the noisy, partially observed $\y$. We show the results of inference with 5 new data observations not included in the dataset in Fig.~\ref{fig:ObsMapExamples}.
To quantify the accuracy of this model we draw 100 independent samples from $p(\z)$ and $p(\fbf)$. We then compute the MSNE for those 100 independent draws. The computed MNSE for $\ubf$ projected on the FE grid is of $5.64 \times 10^{-4}$ and for $\z$ (the $\kappa$(x) field) projected on the FE mesh the MNSE is of $8.35 \times 10^{-3}$.

%--------------------------------------------------------------------------------
\subsection{1D Nonlinear Poisson Problem}\label{sec:OneDPoissonNonlinear}
%--------------------------------------------------------------------------------
%
In this example, we apply a \gls*{PDDLVM} for a 1D nonlinear Poisson problem of the form,
\begin{align}\label{eq:nonLinearStaticPoisson}
   &-\nabla\cdot(\eta(u, x)\nabla u(x)) = f(x),\\
   &\eta(u, x) = \left(S\left|\frac{\md u}{\md x}\right| + 5\kappa(x) \right) / 10,\nonumber
\end{align}
for $x \in (-1,1)$. Here $\z = \{\kappabf, a, b\}$ denotes the coefficients of the $\kappa(x)$ expansion and the boundary condition values respectively and the boundary conditions are $u(-1) = a$, $u(1) = b$. The function $S(x) = {1}/({1+e^{-x}})$ is the sigmoid function. For the current example we parameterize $\kappa(x)$ as a $4^{\text{th}}$ order Chebyshev expansion.
% with a softmax+1 transform.
The solution field is expressed as a $9^{\text{th}}$ order Chebyshev expansion through the $\bpiCH$ mapping.
% that is then projected onto the FE mesh.}
\begin{figure}
        \centering
        \subcaptionbox{FE Solver}{\includegraphics[width=0.45\textwidth]{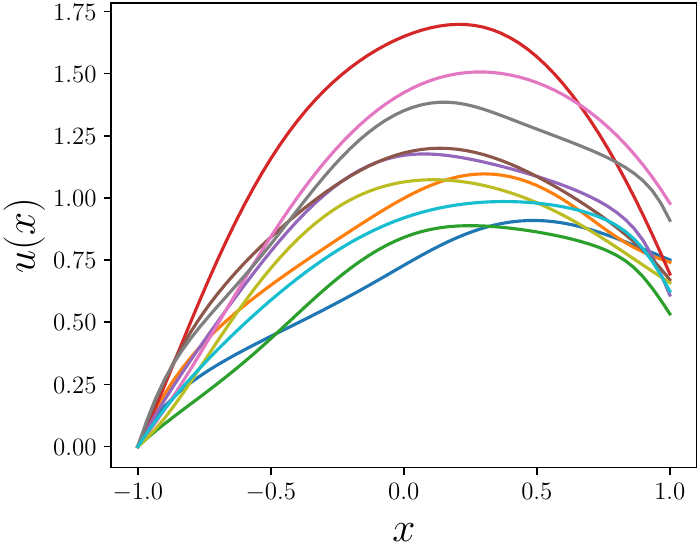}}
        \hspace{3em}
        \subcaptionbox{$q_{\alpha}(\ubf|\z,\fbf)$ }{\includegraphics[width=0.45\textwidth]{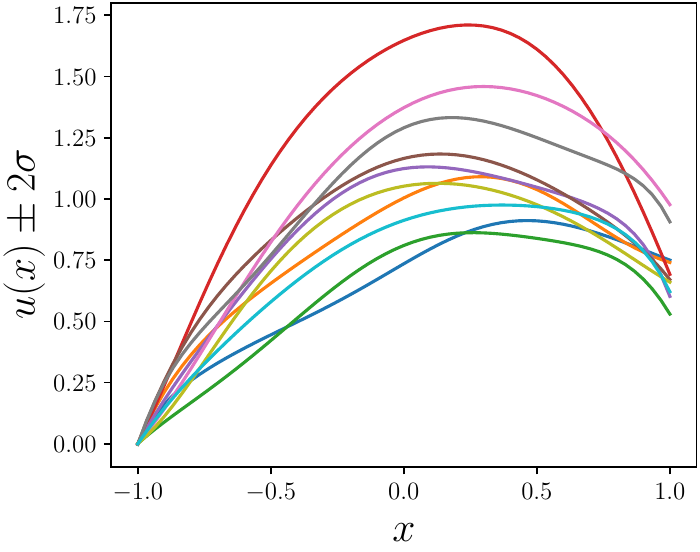}}\\
        \subcaptionbox{True $\kappa(x)$}{\includegraphics[width=0.45\textwidth]{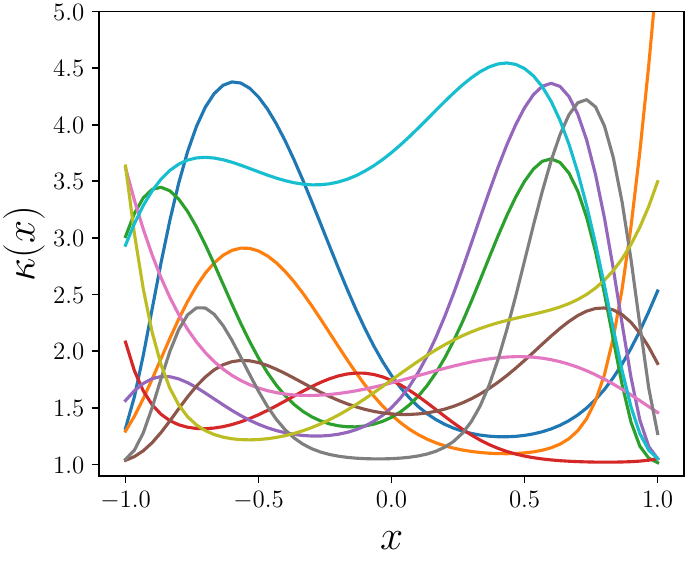}}
        \hspace{3em}
        \subcaptionbox{$p_{\beta}(\z|\ubf, \fbf)$}{\includegraphics[width=0.45\textwidth]{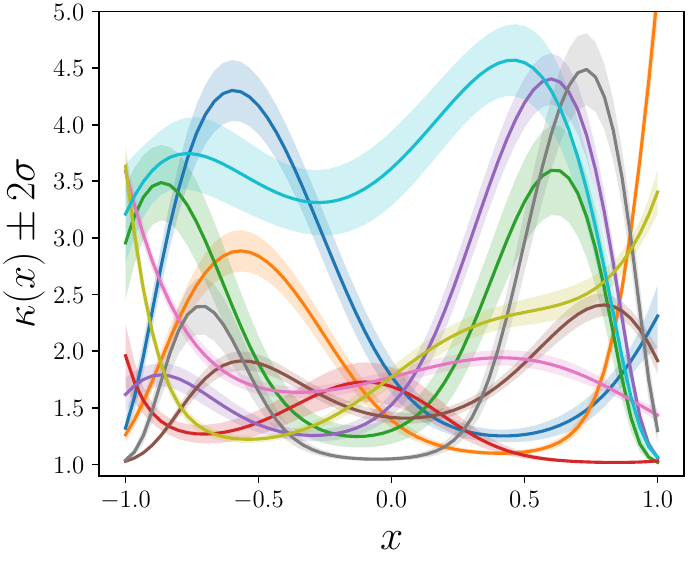}}\\
        \subcaptionbox{$f(x)$}{\includegraphics[width=0.45\textwidth]{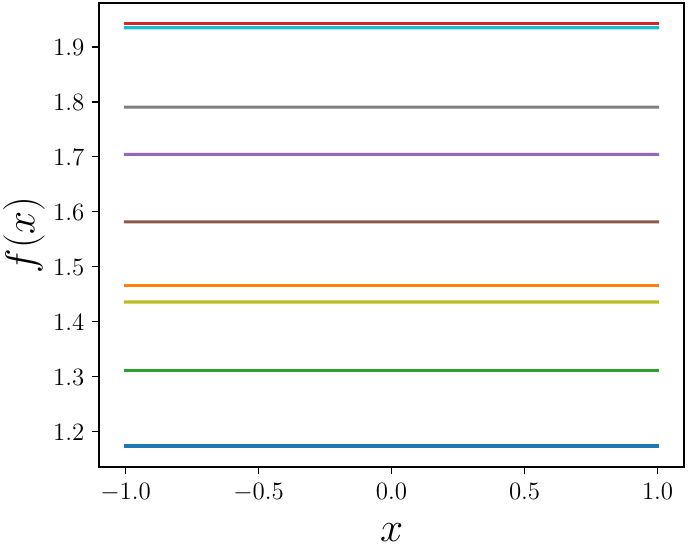}}
    \caption{Results for the 1D nonlinear Poisson example. (a) The true solution for 10 different instances of \{$\z$, $f(x)$\}. (b) The predicted solution with $2\sigma$ interval given by the \gls*{PDDLVM} for those same functions. (c) The 10 true $\kappa(x)$ functions used to compute the solution fields of the first 2 subplots. (d) The \gls*{PDDLVM} reconstruction of $\kappa(x)$. (e) The forcing functions $f(x)$.}
    \label{fig:nonlinearPoisson}
\end{figure}
\begin{table}
\centering
\centering
\begin{tabular}{ c c c }
    \centering
              & $\alpha$-Net & $\beta$-Net \\ 
    \hline
    MNSE     & $6.82\times10^{-4}$ & $2.36\times10^{-3}$ \\   
    \hline
    \% truth in $2\sigma$  & 29.6\% & 87.0\%  \\
    \hline
\end{tabular}
\caption{MNSE for 1D nonlinear Poisson problem and percentage coverage of 2 standard deviations of the output on the ground truth.\\}
\end{table}
 The \textit{approximant} $\bpi_z(\z, x)$ is then only applied on $\kappa(x)$. The priors are: $p(\kappabf_i) = \NPDF(0, 1)$, $p(a) = {\delta}(a)$, $p(b) = \mathcal{U}(0.5, 1)$, and $f(x)$ has a single expansion term drawn from $p(f) = \mathcal{U}(1, 2)$ as is associated to the $0^\text{th}$
Chebysehv polynomial. The NN architecture for this problem is a fully connected 4 layers deep network with 100 neurons per hidden layer. The FE model is discretized with 60 elements. The residual covariance is chosen to be $\bfSigma_{u}=\epsilon_{\r}^2\Ibf$ with $\epsilon_{\r} = 10^{-2}$. The model is trained for $10^6$ iterations with an initial learning rate of $10^{-3}$ that is halved every $2\times 10^5$ iterations with Adam. In Fig.~\ref{fig:nonlinearPoisson}, the predictive distributions as well as the solutions are plotted.

%--------------------------------------------------------------------------------
\subsection{Thin-Walled Flexible Shell}\label{sec:3DshellObject}
\begin{figure}
\centering
\renewcommand{\arraystretch}{2}
\subcaptionbox{}{
\begin{tabular}{ c c c }
              & Forward & Inverse \\ 
    \hline
    MNSE \gls*{PDDLVM}     & 0.00277 & 0.00745 \\   
    \hline
    avg. time \gls*{PDDLVM}  & 0.00346s & 0.0252s  \\ 
    \hline
    avg. time FE  & 10.4s 
\end{tabular}
} \hspace{1cm}
\subcaptionbox{}{
\begin{minipage}{.4\textwidth}
    \includegraphics[width=1\textwidth]{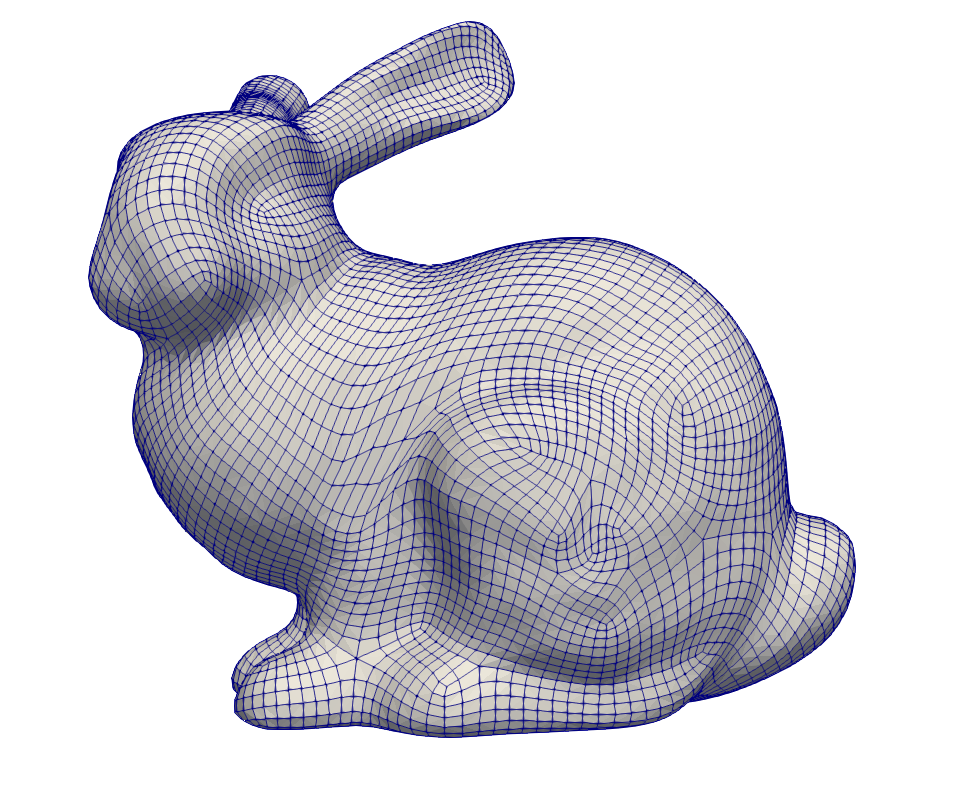}
\end{minipage}
}
\caption{(a) Computational runtime of using the \gls*{FEM}-\gls*{PDDLVM} framework on the thin-walled flexible shell bunny as compared to traditional \gls*{FEM}. (b) The mesh of the bunny geometry used for the finite element discretization, it has 24858 degrees of freedom.}
% \end{table}
\label{tb:resultsPDDLVMBunny}
\end{figure}

%--------------------------------------------------------------------------------
%
In this example, we consider the deformation of the Stanford bunny with an inhomogeneous Young's modulus subjected to self-weight, see Figure~\ref{tb:resultsPDDLVMBunny}.
We analyze this model with an isogeometrically discretized Kirchoff-Love shell finite element scheme \cite{cirak2000subdivision, cirak2011subdivision, febrianto2022digital}. The model is based on the potential energy formulation of a smooth hyper-elastic shell. The weak form of the shell equilibrium equation is obtained by determining the stationary point of the potential energy where $f(x),\, u(x):\bR^3\rightarrow\bR^3$ and the shell is subject to Dirichlet boundary conditions only.

The function $f(x)$ represents the external loading on the shell surface. We choose the forcing to be a uniform vertical pressure representing self-weight. For the Kirchoff-Love model we require basis functions with square integrable second derivatives which are obtained from Catmull-Clark subdivision surfaces \cite{zhang2018subdivision}. In this example, we choose a Dirichlet boundary condition  constraining the displacements at the base of the bunny such that $u(x) = 0, \text{ for } x_2 < b$, where $x_2$ denotes the vertical coordinate axis and $b$ is the chosen height of the base.

To build solution fields with a neural network in a mesh invariant and low dimensional way, we reparameterize the output of the network with a vector valued 3D Chebyshev expansion
\begin{align}
     \hat{\bsu}(\bsx) = \bpiCH(\ubf, \bsx) &= \sum_{l,m,n}\mathbf{U}_{ilmn}T_l(x_{(1)})T_m(x_{(2)})T_n(x_{(3)}),
\end{align}
where $\hat{\bsu}(x) = (\hat{u}_{(1)}(\bsx), \hat{u}_{(2)}(\bsx), \hat{u}_{(3)}(\bsx))^\top$ indexed by $i\in\{1,2,3\}$ and the random vector $\ubf = \text{vect}(\mathbf{U})$ where $\mathbf{U}$ is a four dimensional tensor where the components $\mathbf{U}_{ilmn}$ are Chebyshev coefficients. Furthermore, $\bsx = (x_{(1)},x_{(2)},x_{(3)})^\top \in \Omega \subset \bR^3$.
 % We use a 3D Chebyshev immersion to reduce the number of degrees of freedom used to describe a solution.  
 The vector field $\hat{\bsu}(\bsx)$ denotes the displacement of the bunny surface and is computed on the intersection of the volume and the surface. Describing a 3 component vector field in 3 dimensions this way requires $3N^3$ Chebyshev coefficients, where $N$ denotes the number of basis function per dimension. We choose a sixth order Chebyshev expansion with 7 coefficients per dimension so that we have 1029 Chebyshev expansion coefficients. On the other hand, the quadrilateral Stanford bunny mesh has 8286 nodes and 24858 degrees of freedom. The bunny geometry width, height and length are 154.8, 151.9, and 118.6 respectively. All coordinates plus a $10$ unit buffer on all sides are then mapped to the unit cube for the centered Chebyshev reparametrization. The shell has a thickness of $1$, a vertical volumetric pressure field of $10^{-5}$, a Poisson ratio of 0.3, and the height of the fixed base is of $7$. For this example, the physically interpretable latent variable $\z$ is the set of parameters characterizing $\kappa(x_{(2)})$ which varies across the height of the shell to mimic the control of the material distribution that one might have in 3D printing contexts. It represents the modulus of elasticity expressed as a second order Chebyshev expansion passed through a Softplus function and multiplied by $10^9$. 

\begin{figure}
        %\rulesep
        \centering
        \subcaptionbox{Forward \gls*{FEM}}{\includegraphics[width=0.3\textwidth]{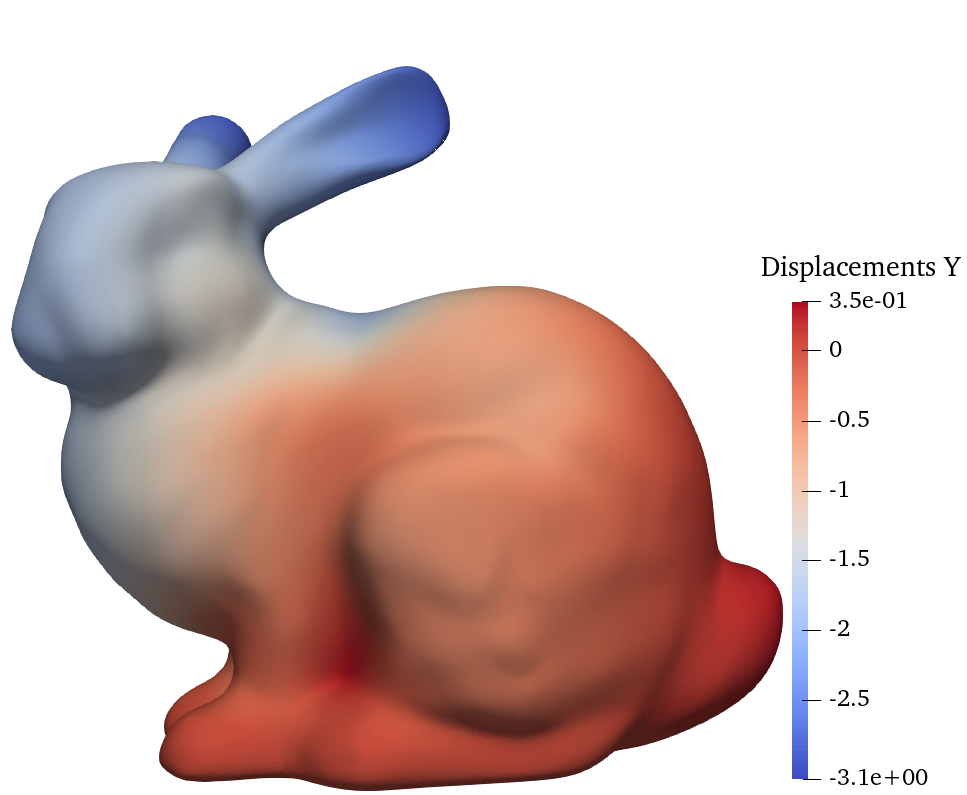}}
        \hspace{6em}
        \subcaptionbox{Mean Forward \gls*{PDDLVM} }{\includegraphics[width=0.3\textwidth]{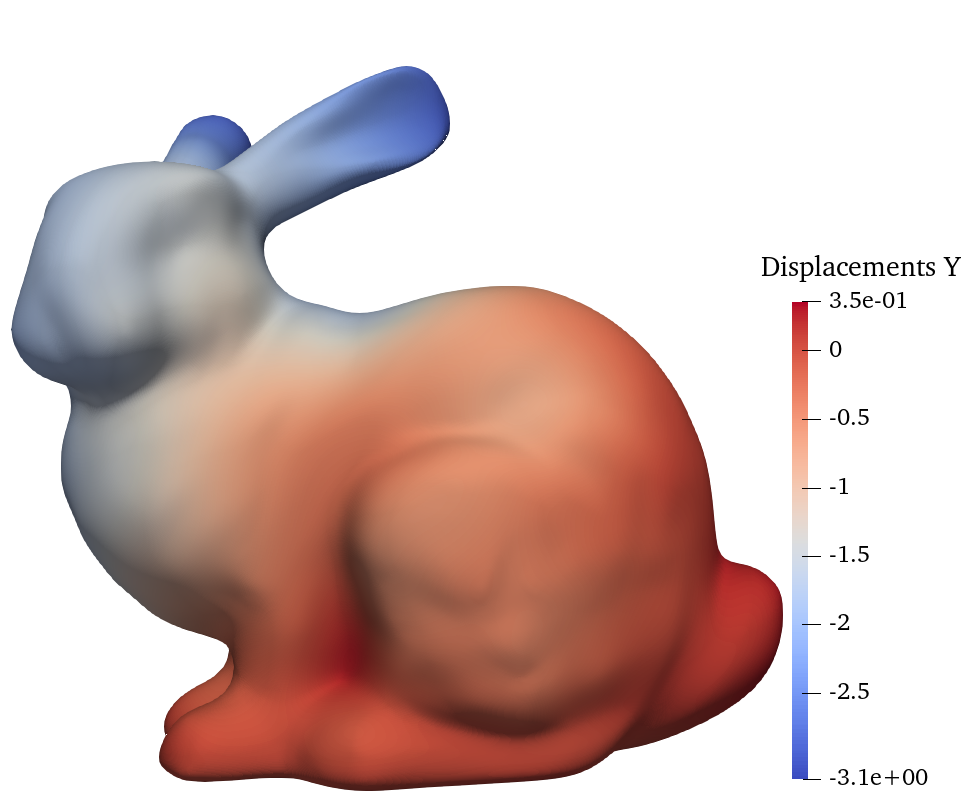}}
        \\
        \subcaptionbox{Abs. Error (forward) }{\includegraphics[width=0.3\textwidth]{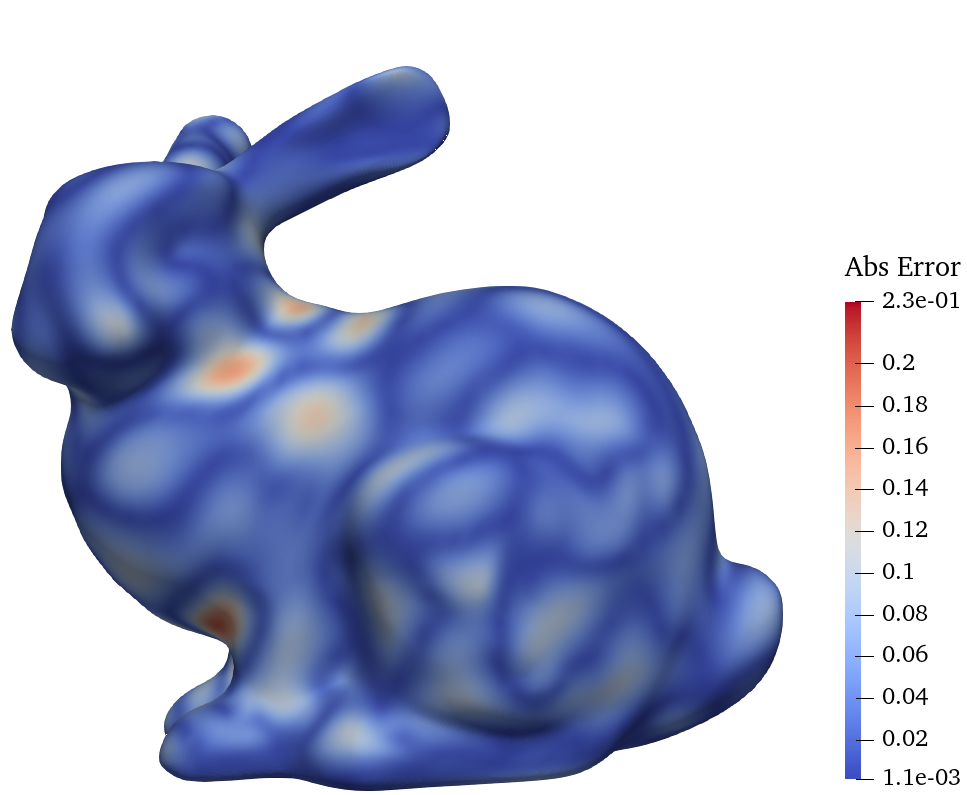}}
        \hspace{6em}
        \subcaptionbox{St. Dev. (forward) }{\includegraphics[width=0.3\textwidth]{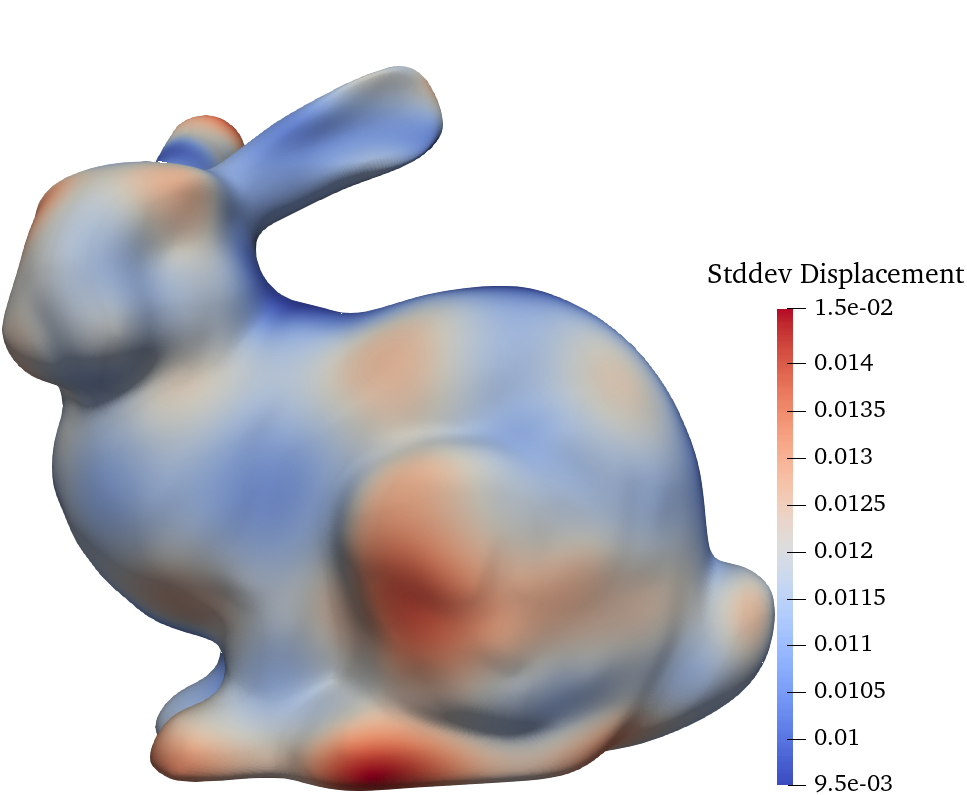}}
        \\
        \subcaptionbox{True Parameter Field}{\includegraphics[width=0.3\textwidth]{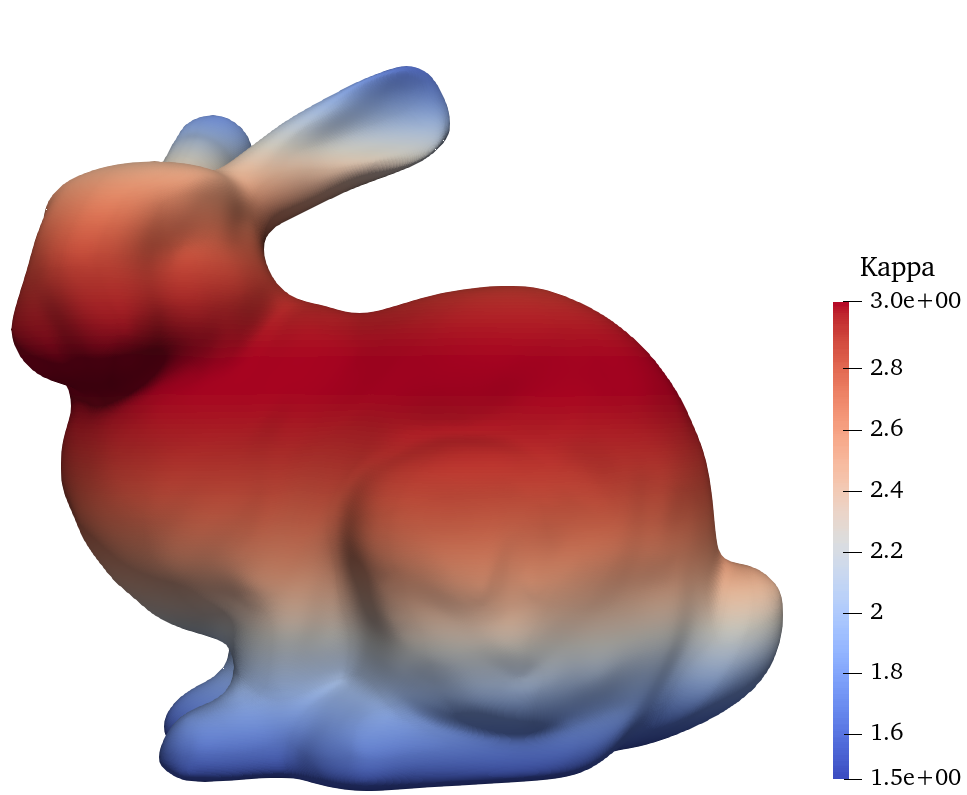}}
        \hspace{6em}
        \subcaptionbox{Mean Inverse \gls*{PDDLVM} }{\includegraphics[width=0.3\textwidth]{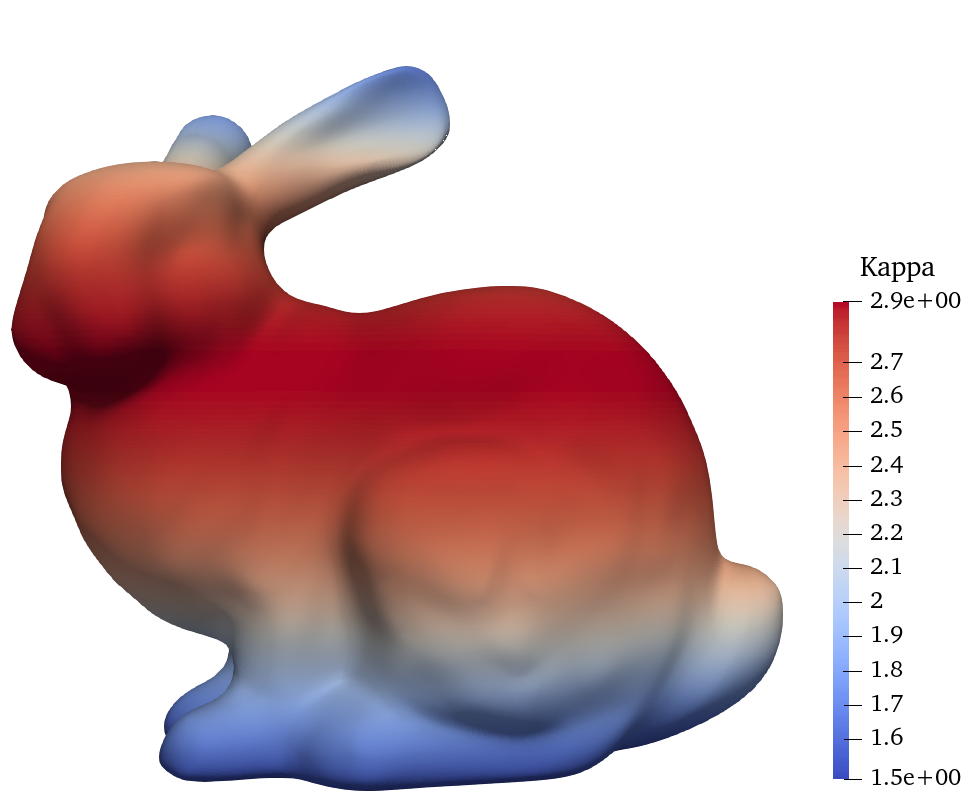}}
        \\
        \subcaptionbox{Abs. Error (inverse)}{\includegraphics[width=0.3\textwidth]{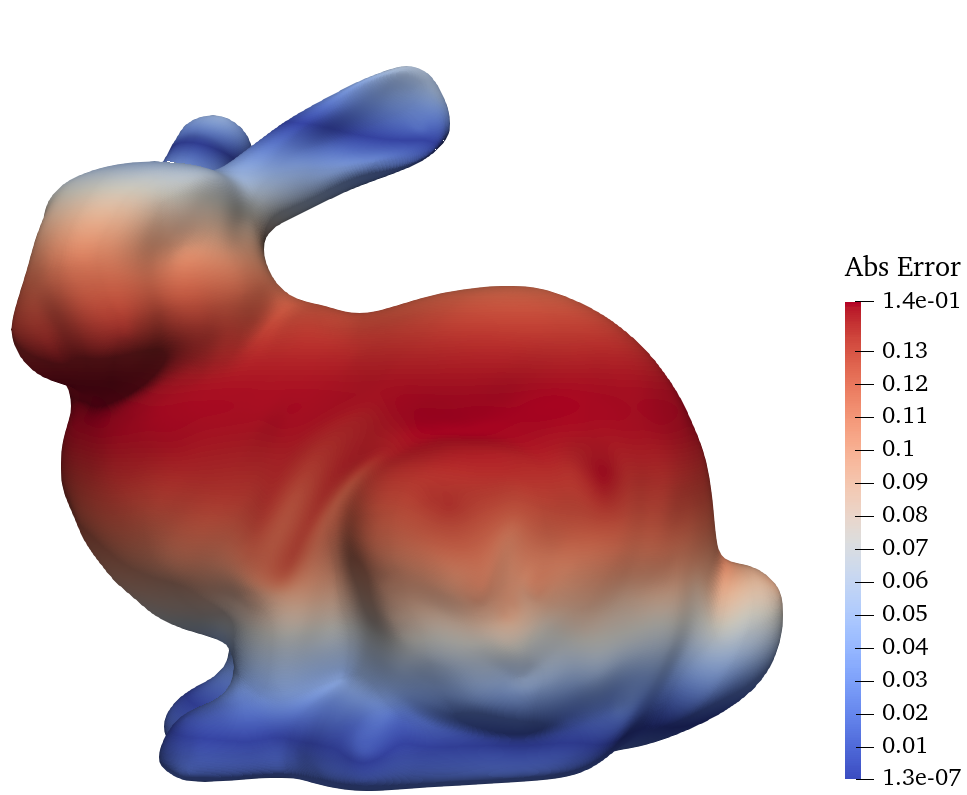}}
        \hspace{6em}
        \subcaptionbox{St. Dev. (inverse)}{\includegraphics[width=0.3\textwidth]{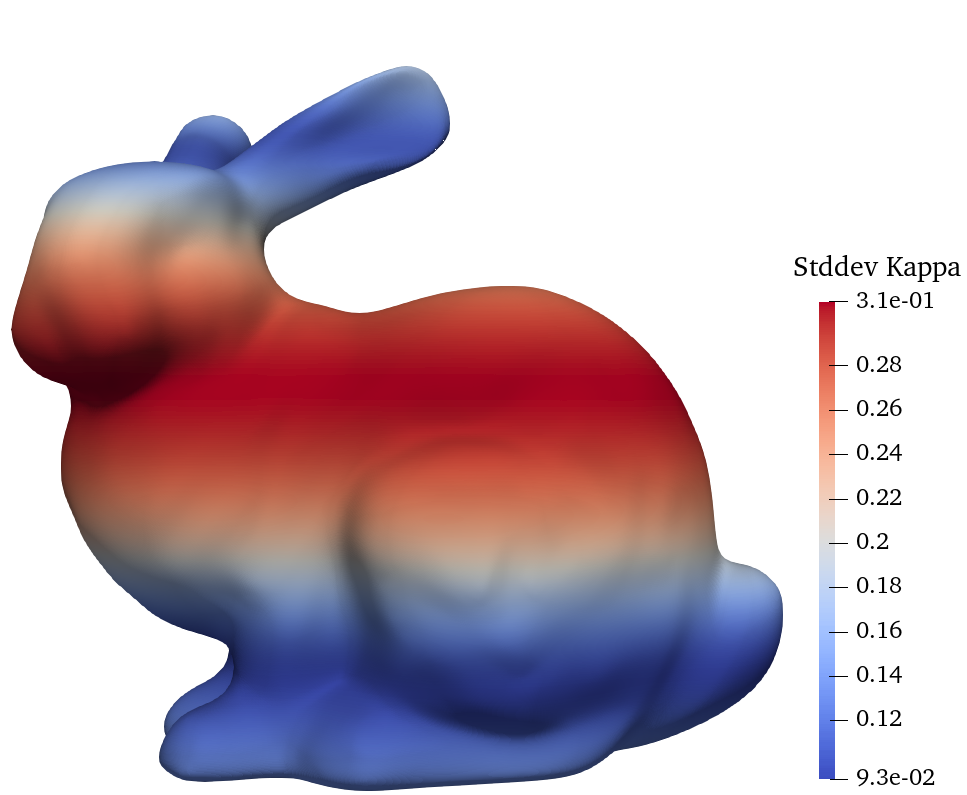}}
   \caption{Results from a trained \gls*{FEM}-\gls*{PDDLVM} of a thin-walled 3D shell bunny. The top row shows (a) the displacement field obtained with a forward \gls*{FEM} solution, (b) mean estimate provided by \gls*{PDDLVM}, (c) absolute error between (a) and (b), (d) standard deviation. In the bottom row, similarly, we have (e) true parameter field, (f) mean estimate of the parameters provided by the inverse of \gls*{PDDLVM}, (g) absolute error, (h) standard deviation. The agreement between (g) and (h) demonstrates that when the difference between our method and the true parameter field is larger, our uncertainty estimates are coherent and demonstrates the regions with larger error accurately. See Sec.~\ref{sec:3DshellObject} for more details.}
   \label{fig:BunnyPDDLVM}
\end{figure}

With the \gls*{FEM} formulation we can take advantage of well-known preconditioning techniques \cite{grote1997parallel} for iterative solvers. Employing a simple preconditioner can greatly accelerate and stabilize convergence. This preconditioning is formulated as a reparametrization of the residual with the stiffness matrix of a representative sample $\z$ of $p(\z)$. To do this, we pre-multiply the residual vector $\r$ by ${\A}_{\bar{\z}}$. The matrix ${\A}_{\bar{\z}}$ is the stiffness matrix assembled with $\bar{\z}$ which the mean of $p(\z)$. This yields the preconditioned residual formulation as ${\A}_{\bar{\z}}\r$ which was found to greatly accelerate convergence and improve stability of the optimization routine. Both $\alpha$ and $\beta$ networks are fully connected neural networks with 3 hidden layers of 2.5k neurons with ``swish'' activation functions \cite{ramachandran2017searching}. The \gls*{PDDLVM} was trained for 50k iterations for a total of 16.6 hours. Once trained, we draw new samples of $\z$ from its prior, we then use both the \gls*{FEM} model and the emulator to obtain the solution fields for comparison. Fig.~\ref{fig:BunnyPDDLVM} shows the predictions of the trained \gls*{PDDLVM}. The top row pertains to the forward operation, the bottom row focuses on the inverse mapping. We plot the \gls*{FEM} solutions, the predictive mean, the standard deviation, as well as the absolute error. Table \ref{tb:resultsPDDLVMBunny}(a) summarizes the mean normalized squared errors of the means of both the forward and inverse maps as well as the average time taken per sample for 100 independent draws from the prior $p(\z)$.

 When computing the residual term necessary for \gls*{FEM}-\gls*{PDDLVM}, we need to update the stiffness matrix for new samples of the PDE parameters $\z$. In the governing equation, the domain integral depends linearly on Young's modulus and we take its value to be constant across each element. Thus, when we compute the residual for a new sample of $\z$ we do not need to recompute the quadrature for the integration of the r.h.s. of the weak form. We can instead stash the element stiffness matrices for a Young's modulus of 1 and post-multiply the already integrated element matrices with the new values for the Young's modulus at the centroid of the element. From this we can then assemble the new matrix, bypassing the expensive quadrature routine. This observation is important when computing the residual tens of thousands of times for large problems. However, if we are interested in marginalizing over a quantity that cannot be factorized out of the element stiffness integrals then the quadrature must be computed every iteration. 

% \subsection{\gls*{PINN}-\gls*{PDDLVM}}\label{sec:PINN_PDDLVM}
%--------------------------------------------------------------------------------
\subsection{ 1D Nonlinear Heat Equation}\label{sec:PINN_PDDLVM}
%--------------------------------------------------------------------------------
%
\begin{figure}
    \centering
    \subcaptionbox{$\bmu_{\alpha}(x,t)$}{\includegraphics[trim=0 10 0 0,clip, width=0.49\textwidth]{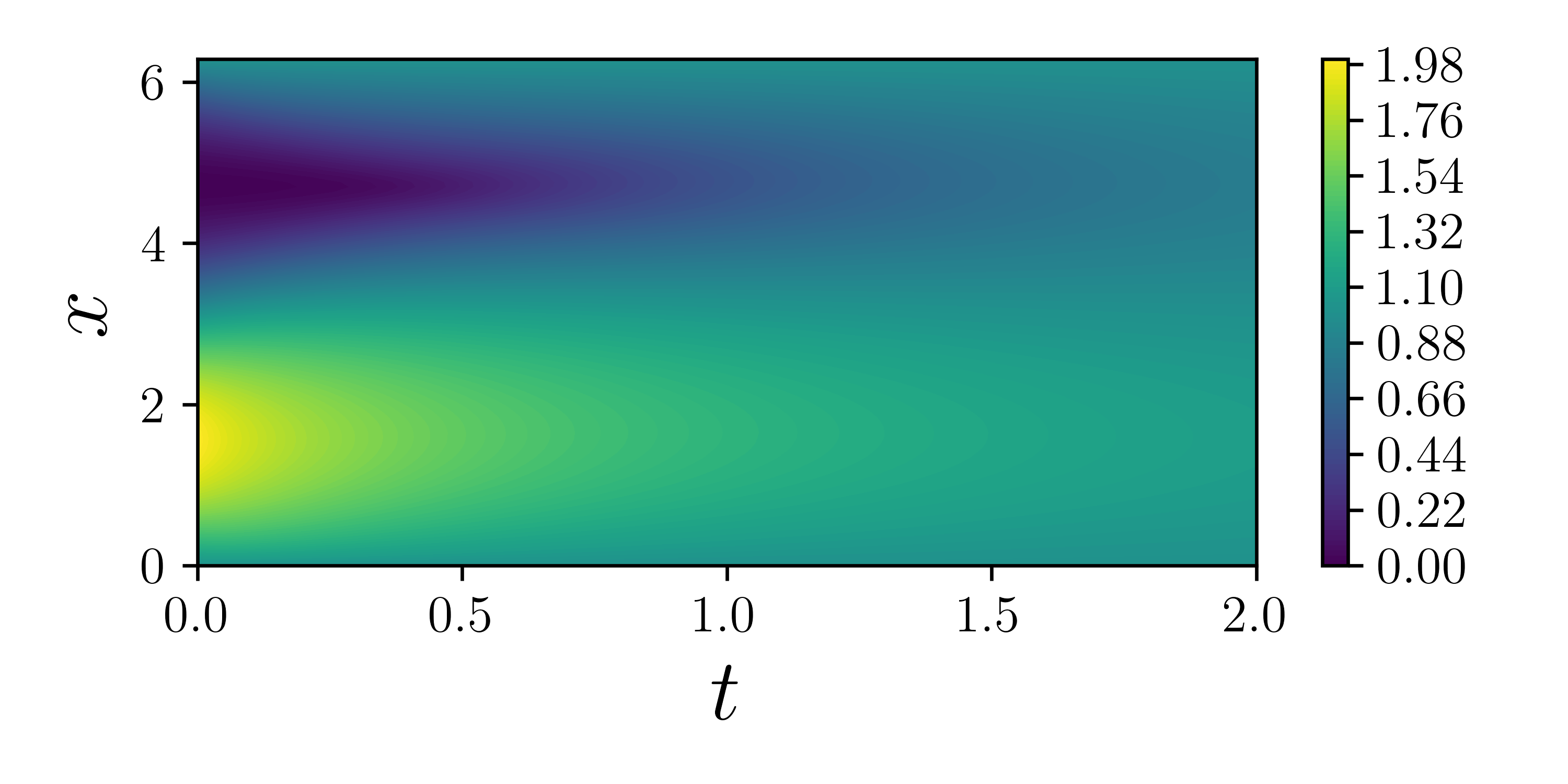}}
    \subcaptionbox{$2\sigma_{\alpha}(x,t)$}{\includegraphics[trim=0 10 0 0,clip, width=0.49\textwidth]{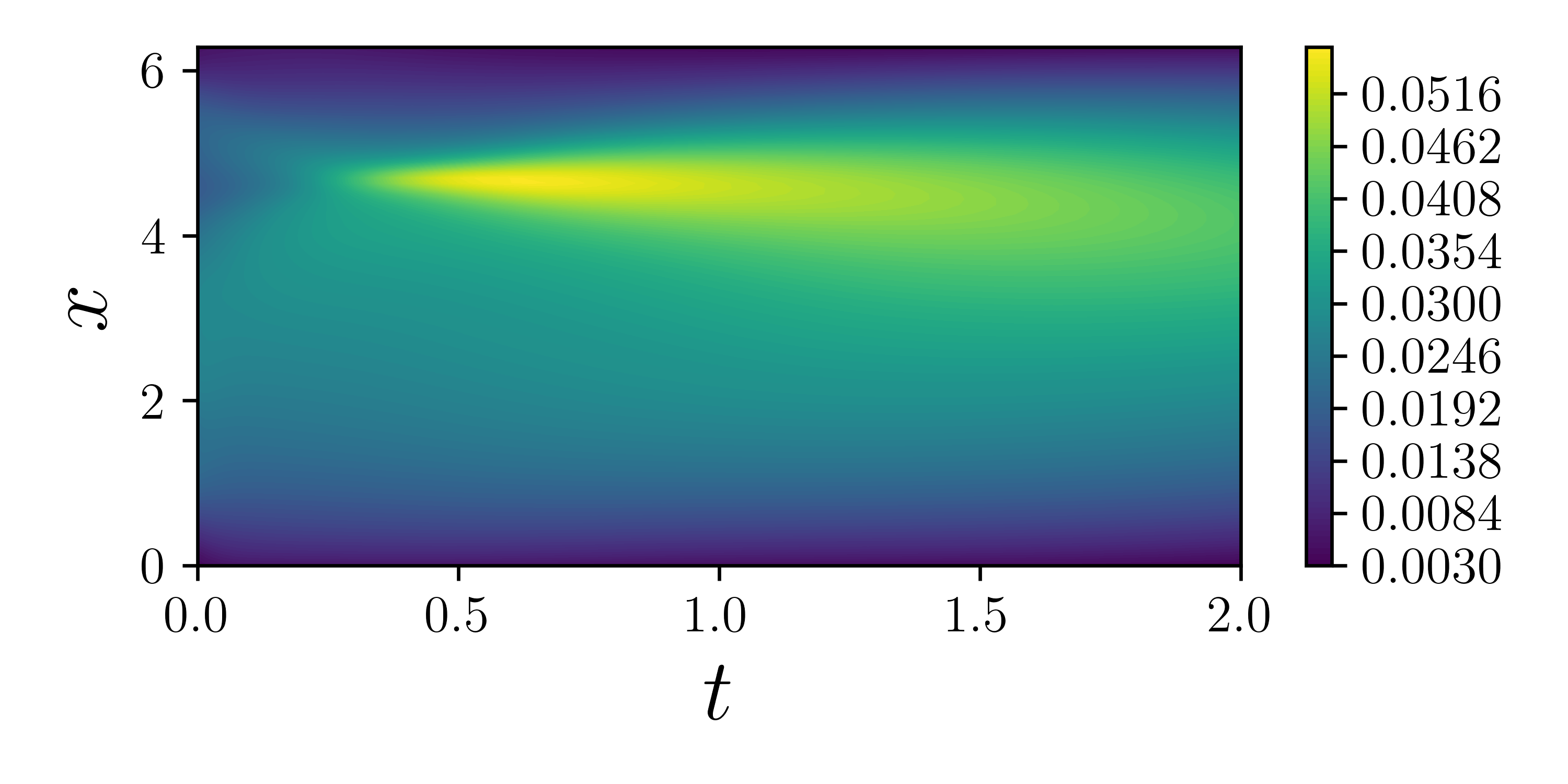}}
     \caption{The distribution $q_{\alpha}(\ubf|\z, \fbf)$ for a parameter sample $\gamma, \kappa$ for the time dependent nonlinear heat equation as defined in \eqref{eqn:nonLinearTimeDptPoisson} and discretized with \gls*{PINN}-\gls*{PDDLVM}. In (a) we show the mean of the predictive field and in (b) we show the 2 standard deviation of the predictive solution field. Notice the uncertainty is lowest at the fixed boundaries at $x=0, \; x=2\pi$ and highest in the region of higher non-linearity.}
     \label{fig:samplePredictionPINNPDDLVM}
\end{figure}

To showcase the modularity of the \gls*{PDDLVM} framework, we consider next a time dependent nonlinear heat equation discretized with a \gls*{PINN}\cite{raissi2019physics}. We can treat this as a spatio-temporal problem where we work with spatial and temporal dimensions the same way~\cite{raissi2019physics}. When combined with the \gls*{PINN} framework, our solution field predictive distribution and the inverse distribution are formulated in a general way as
\begin{align}
    q_{\alpha}(\ubf|\z, \fbf) &= \NPDF\left(\bmu_{\alpha}(x_{1:n}, t_{1:n}, \z, \fbf), \text{diag}(\sigma^2_{\alpha}(x_{1:n},t_{1:n},\z,\fbf)\right),\nonumber\\
    p_{\beta}(\z|\ubf, \fbf) &= \NPDF(\bmu_\beta(\ubf), \bSigma_{\beta}(\ubf)),
\end{align}
% \begin{wraptable}{r}{11cm}
\begin{table}[h]
\centering
\begin{tabular}{ c c c }
    \centering
              & PINN \& Conv-Net & PINN-PDDLVM \\ 
    \hline
    Mean Residual Forward    & $2.83 \times 10^{-4}$ & $1.23 \times 10^{-4}$ \\   
    \hline
    MNSE Inverse  & $6.35\times10^{-3}$ & $2.83\times10^{-3}$   \\
    \hline
    \% truth in $2\sigma_{\beta}$ &     n/a              & 98.06\%\\
    \hline
    total training time &  6440s      & 6808s

\end{tabular}
\caption{Comparison of PINN-PDDLVM model and a coupled PINN and convolutional inversion network.  For details on the \gls*{PINN} residual formulations, see \eqref{eq:pinnResHeat}.}
\end{table}
% \end{wraptable}
where $\ubf \in \bR^{n}$ and are jointly trained as per the framework in Sec.~\ref{sec:prob_model:dataless}. The samples $\ubf$ are then given to the various other probability density functions as well as the residual evaluator in \eqref{eq:discreteDiffOp}. Here the collocation method with $w_i(x, t) = \delta(x-x_i) \delta(t-t_i)$ is used reducing the integration to a point evaluation of the PDE on the grid. We use automatic differentiation through variable $\ubf$ and the neural network to obtain $\partial_x \hat{u}(x)$ with the reparametrization trick. The \texttt{parameter-to-solution} map takes in a pair of $x,t$ coordinates along with the PDE parameters. To obtain the distribution across the domain we pass in a grid of $x,t$ coordinates. For the inverse map, we pass a picture of the entire field as a grid of evaluations. A convolutional structure is well adapted for dealing with this kind of spatial data.

The PDE we analyze is the heat equation with a temperature dependent conductivity function. We specify the second order nonlinear parametric time dependent PDE as 
\begin{equation}
    \frac{\partial u(x,t)}{\partial t} = \frac{1}{\gamma}\nabla\cdot(\eta(u, \kappa)\nabla u(x, t)), \quad \quad \eta(u, \kappa) = \left|\frac{u(x,t)\kappa^2+1}{\kappa}\right|,
    \label{eqn:nonLinearTimeDptPoisson}
\end{equation}
 for $x \in (0,2\pi)$, with initial conditions and boundary conditions $u(0, t) = 1$,  $u(2\pi, t) = 1$, and $u(x, 0) = \sin(x) + 1$. The $\eta(u, \kappa)$ function represents a material whose conductance increases linearly as the temperature increases with a slope and intercept are governed by the scalar $\kappa$.
We construct $\bfSigma_{\r} = \text{diag}\left([\epsilon_{D_0}^2,\hdots,\epsilon_{D_d}^2, \epsilon_{B_0}^2, \hdots,\epsilon_{B_b}^2, \epsilon_{I_0}^2, \hdots,\epsilon_{I_i}^2]\right)$ where each $\epsilon$ denotes the confidence for the $d$ domain nodes, the $b$ boundary nodes, and the $i$ initial condition nodes respectively. We choose a residual standard deviation $\epsilon_D = 0.01$, while to strongly enforce the initial and boundary conditions, we choose $\epsilon_B=0.001$ and $\epsilon_I=0.001$.  The PDE is also parameterized with $\gamma$ which denotes the heat capacity of the material. We train the \gls*{PDDLVM} with marginalization over the prior $p(\z) = \mathcal{U}([1,5]\times[1,5])$ where $\z = \{\kappa, \gamma\}$.
 \begin{figure}[h!]
    \centering
    \subcaptionbox{$\log R_{\text{PINN}}(\gamma, \kappa)$ }{\includegraphics[width=0.45\textwidth]{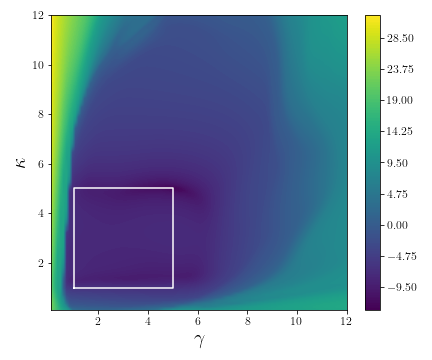}}
    \hspace{1em}
    \subcaptionbox{$\log R_{\mu_{\alpha}}(\gamma, \kappa)$ }{\includegraphics[width=0.45\textwidth]{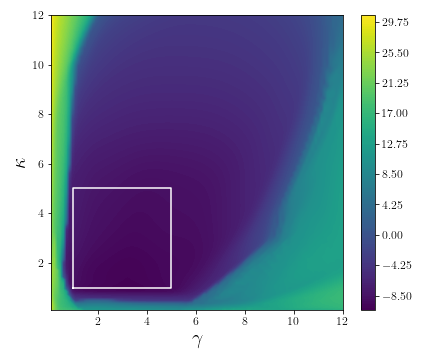}}
    \\
    \subcaptionbox{$\log \bar{\sigma}_{\alpha}(\gamma, \kappa)$ }{\includegraphics[width=0.45\textwidth]{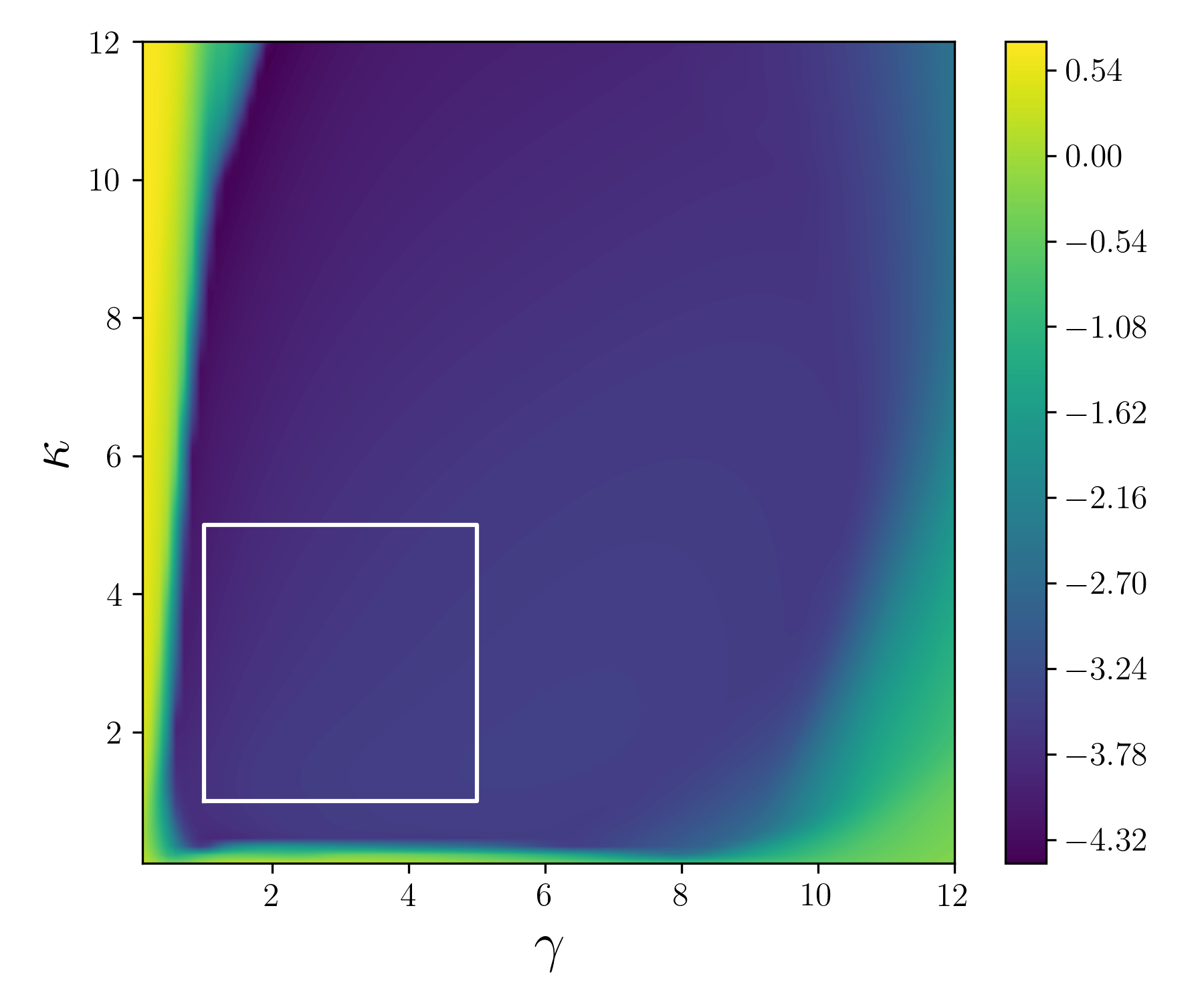}}
    \caption{A comparison of (a) \gls*{PINN} and (b) \gls*{PINN}-\gls*{PDDLVM} residuals and in  (c) the average \gls*{PINN}-\gls*{PDDLVM} standard deviation. The quantities are computed over a 2-dimensional range of values of parameters $\gamma$ and $\kappa$. The white square denotes the bounds for the observed $\kappa, \gamma$ during training as well as the bounds over which the mean residual is computed and is defined by the prior $p(\z) = p(\gamma)p(\kappa) = \mathcal{U}([1,5])\times\mathcal{U}([1,5]) $.}
    \label{fig:residualsPINNPDDLVM}
\end{figure}

 In Fig.~\ref{fig:residualsPINNPDDLVM} we plot the $\log$ \gls*{PINN} residual for a range of PDE parameters $\kappa$ and $\gamma$ and we plot the $\log$ mean standard deviation of the $\alpha$ network. In Fig.~\ref{fig:samplePredictionPINNPDDLVM} we show a predicted distribution over the solution field of the heat equation for a sample of $\gamma, \kappa$ drawn from the prior and computed through the \gls*{PINN}-\gls*{PDDLVM} framework.

The \texttt{parameter-to-solution} network for both the standard \gls*{PINN} and \gls*{PINN}-\gls*{PDDLVM} models have 4 fully connected hidden layers of 100 neurons with ``swish'' activation functions. For the PINNs results we sampled 2.5k combinations of $\gamma$ and $\kappa$ between $[1., 5.] \times [1., 5.]$ and we trained a PINN-network for $5\times10^5$ iterations. We then used this trained PINN to output solutions fields and create a supervised training dataset for the convolutional inverse network. Both the deterministic inversion network (Conv-Net) and PINN-PDDLVM-$\beta$ networks are 2D convolutional neural networks with 3 layers of 2D convolutions with swish activation, kernel size of 4, stride (2,2) and [8, 16, 32] filters followed by a final dense layer of 500 neurons ending with either 2 or 4 output neurons for either the Conv-Net network or the PINN-PDDLVM-$\beta$ networks respectively. 

We train the Conv-Net network with a squared error loss, minibatched one example at a time with Adam and a learning rate of $10^{-3}$. When computing the residual for comparison we employ the standard \gls*{PINN}-style \cite{raissi2019physics}  residual as 
\begin{align}
    R_h(\gamma, \kappa) &= \frac{1}{NM}\sum_i^N \sum_j^M\left(  -\frac{\partial h(x_i,t_j)}{\partial t} + \frac{1}{\gamma}\nabla\cdot(\eta(h(x_i,t_j), \kappa)\nabla h(x_i,t_j))\right)^2 \nonumber\\&+ \frac{1}{n}\sum_k^n\left(h(x_k, 0) - u(x_k, 0)\right)^2 + \frac{1}{m}\sum_l^m\left[\left(h(0, t_0) - u(0, t_0)\right)^2 + \left(h(2\pi, t_0) - u(2\pi, t_0)\right)^2 \right],
    \label{eq:pinnResHeat}
\end{align} 
where the second and the third terms are the residuals for the initial and the boundary conditions. It is the $\log$ of \eqref{eq:pinnResHeat} that is plotted in Figs.~\ref{fig:residualsPINNPDDLVM} but it is not the residual used for \textit{training} the \gls*{PINN}-\gls*{PDDLVM}, for that we use \eqref{eq:discreteDiffOp}.

%--------------------------------------------------------------------------------
\subsection{Inhomogeneous Wave Equation}\label{sec:wave}
%--------------------------------------------------------------------------------
\begin{figure}[h]
        \centering
        \subcaptionbox{True wave A}{\includegraphics[width=0.45\textwidth]{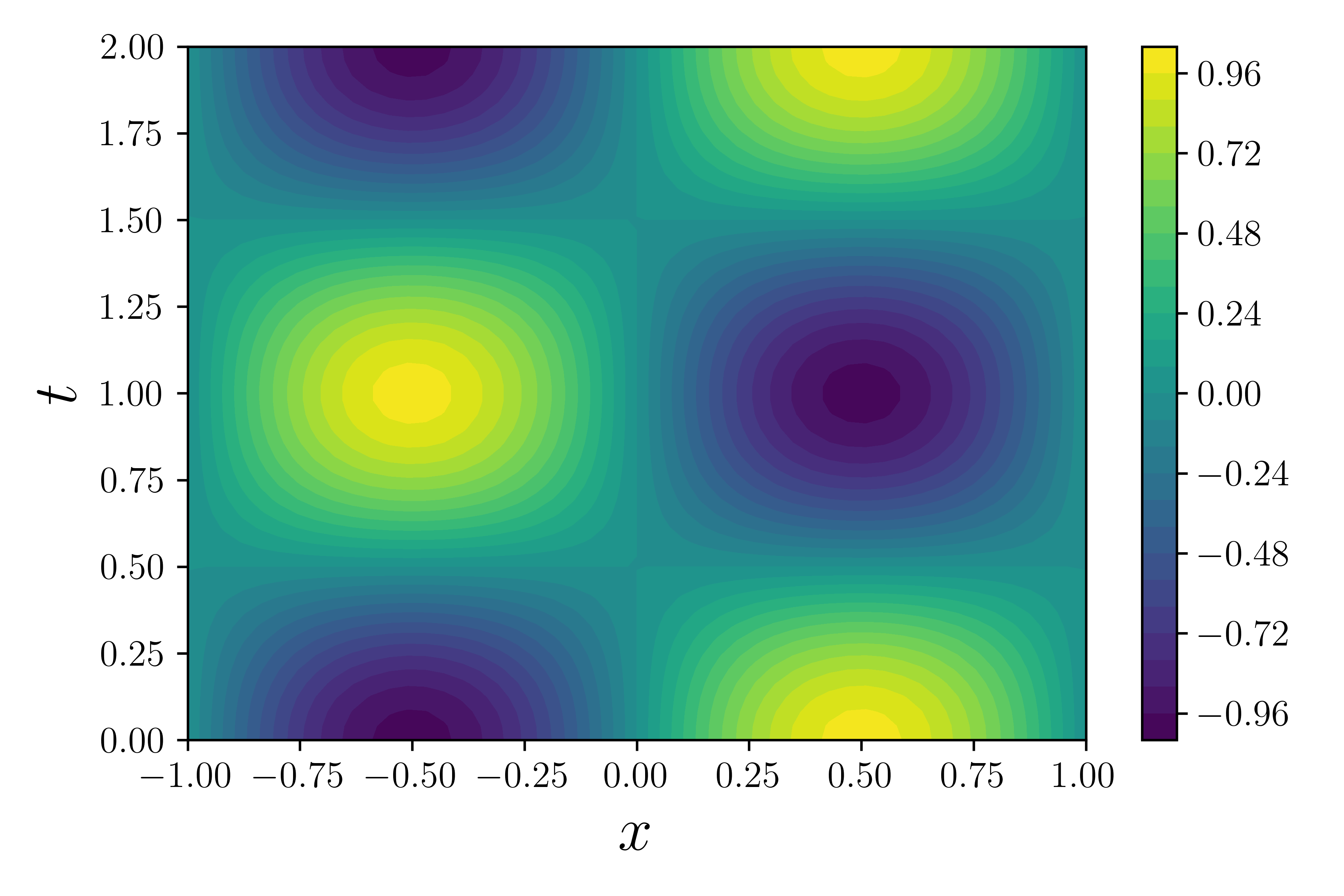}}
        \hspace{2em}
        \subcaptionbox{$\alpha$-Net Mean A}{\includegraphics[width=0.45\textwidth]{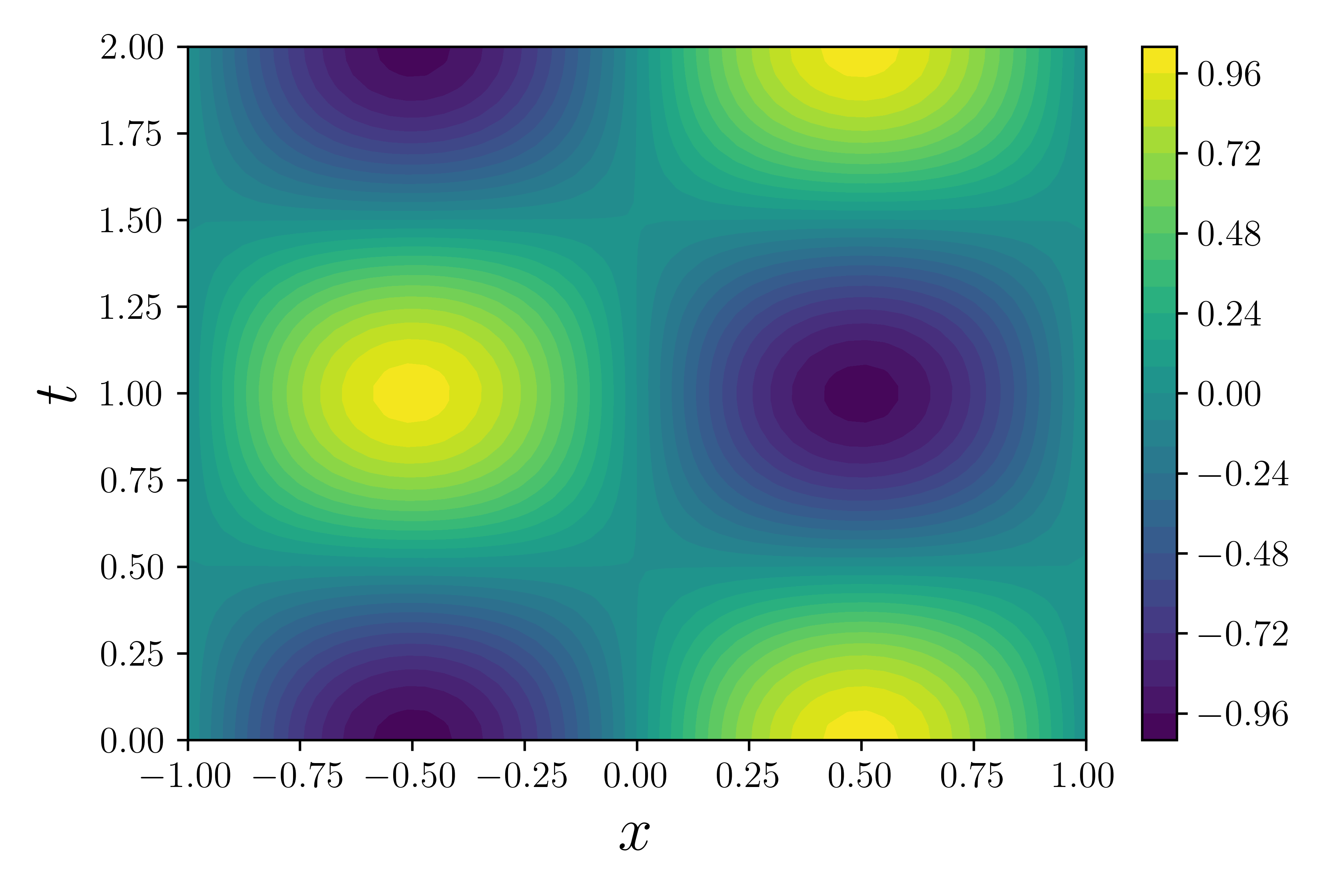}}
        \\
        \subcaptionbox{True wave B}{\includegraphics[width=0.45\textwidth]{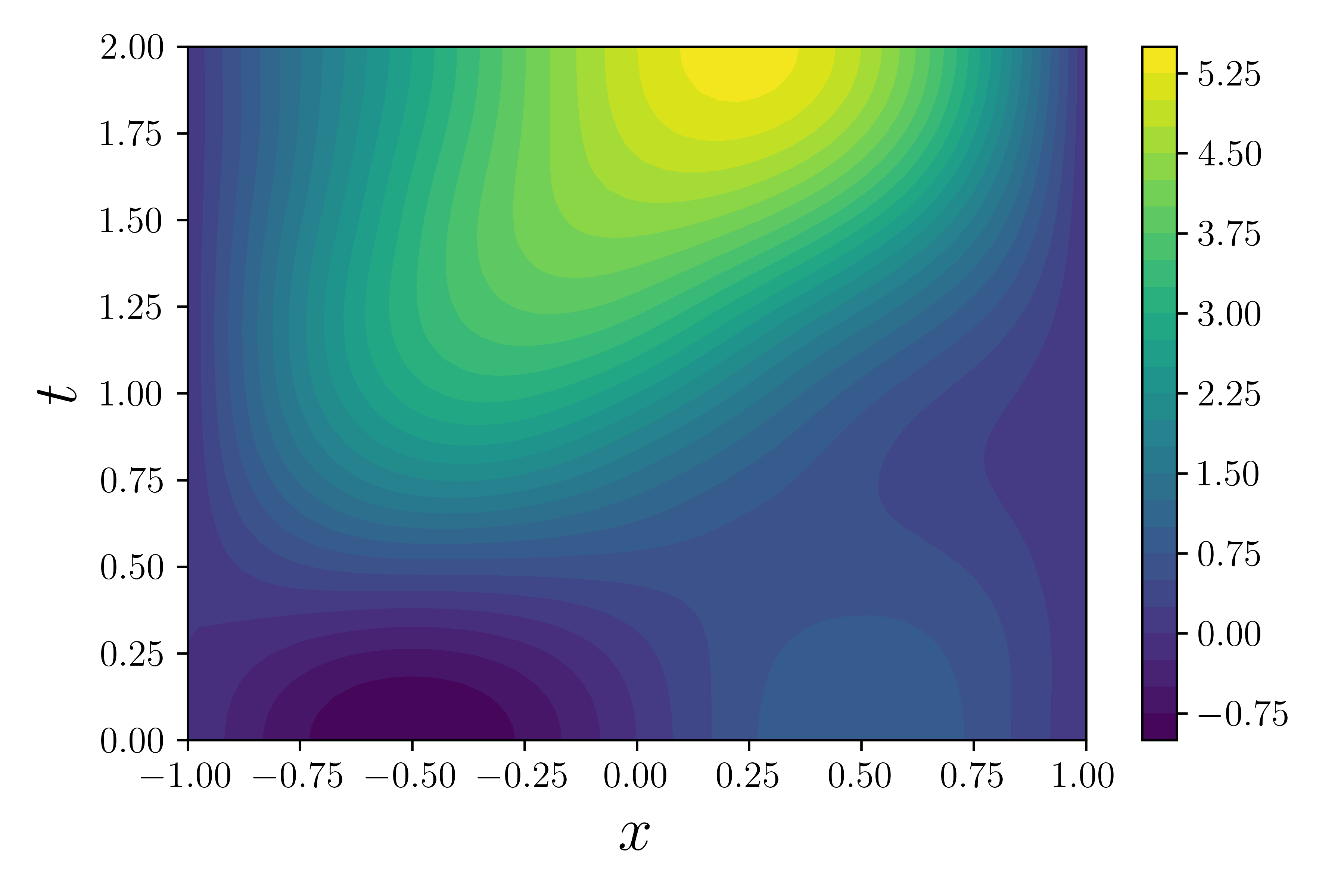}}
        \hspace{2em}
        \subcaptionbox{$\alpha$-Net Mean B}{\includegraphics[width=0.45\textwidth]{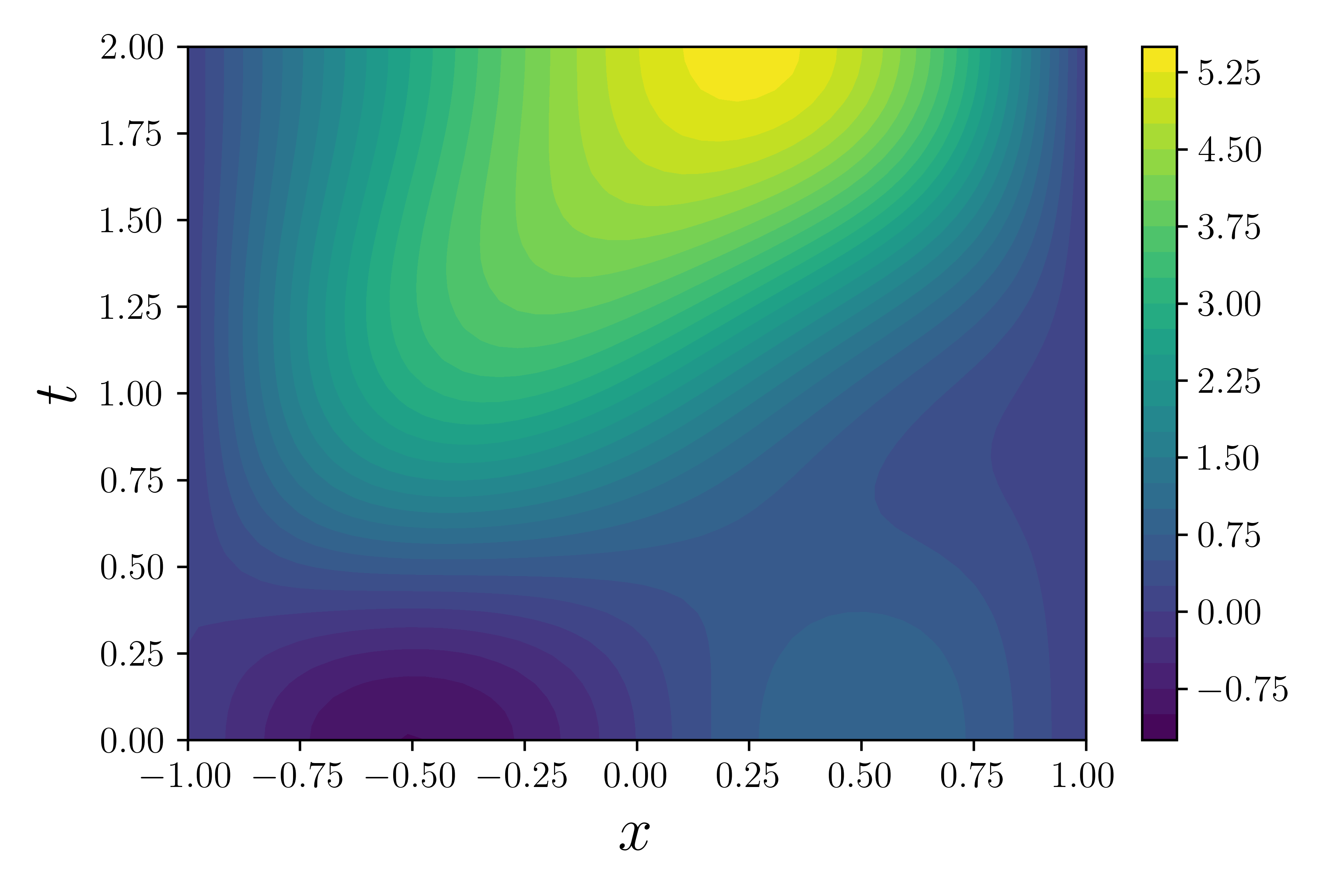}}
    \caption{Examples of ground truth and mean predictions from the $\alpha$-Net of a \gls*{PINN}-\gls*{PDDLVM} for 2 sets of waves with different forcing. Wave A: $f = 0$, Wave B: $f=-5x^3$.}
    \label{fig:inhomogenousWaveEq}
\end{figure}
Here, we compare the use of a 2D Fourier Neural Operator (FNO) trained on a supervised data of PDE solutions with an unsupervised PINN-PDDLVM model. The PDE in question is the inhomogeneous wave equation with parameterized forcing,
\begin{align}\label{eqn:inhomogenousWave}
    &\frac{\partial^2 u}{\partial t^2} - \frac{\partial^2 u}{\partial x^2} = f(x),\\
    &f(x) = \sum_i f_i x^i.\nonumber
\end{align}
Here $f_i$ make up the vector $\fbf$ and are the parameters which make up the polynomial expansion representation of the forcing. In the current example we use 4 terms in the forcing expansion where all components are drawn from $\mathcal{U}(-5, 5)$. The solutions for FNO datasets and the validation set are assembled and computed from samples of $\fbf$ drawn from this distribution. 
\begin{table}[h!]
\centering
\begin{tabular}{ c c c c c c   }
    \centering
                  & FNO-50 & FNO-100   & FNO-500 & FNO-1000 & PDDLVM \\ 
    \hline
    MNSE-Forward & $1.78\times10^{-2}$ & $6.20\times10^{-4}$ & $1.18\times10^{-4}$  &  $9.57\times10^{-5}$ & $1.29 \times 10^{-4}$ \\   
    \hline                                  
    MNSE-Inverse  & $9.40\times10^{-3}$  & $8.67\times10^{-4}$&   $1.69\times10^{-4}$ & $1.31\times10^{-4}$ & $6.51\times10^{-4}$   \\
\end{tabular}
\caption{Comparison of MNSE for supervised FNOs with differing number of data points (number following FNO) against an unsupervised PINN-PDDLVM.}\label{table:FNO_results}
\end{table}
In Fig.~\ref{fig:inhomogenousWaveEq} we show 2 example ground truth solutions from the dataset as well as the mean predictions of those two example given by the \gls*{PINN}-\gls*{PDDLVM}. The FNOs for the forward and inverse directions have 4 hidden layers with 32 channels and 12 Fourier modes. We train the FNOs for $5\times10^4$ iterations for different number of training data. The training data is obtained from a Chebyshev based solver of $21\times21$ basis functions. In Table~\ref{table:FNO_results} we compare this with an unsupervised PINN-PDDLVM trained for $2.5\times10^5$ iterations with a similar architecture to that of Sec.~\ref{sec:PINN_PDDLVM}.

%--------------------------------------------------------------------------------
\section{Conclusions}
%--------------------------------------------------------------------------------
%
\gls*{PDDLVM} is a fully probabilistic generative model for parametric PDEs that builds an interpretable latent space representation and provides uncertainty estimates on the predictions. The Bayesian view allows us to construct a collection of probabilistic mappings that are connected together in a sound and principled way. The ability to produce forward and inverse solutions with uncertainty estimates is of significant importance to the practical use of such methods. It is particularly crucial when dealing with inverse problems  as they are inherently ill-posed.

The modular nature of the developed framework allows for an easy transition between meshless (\emph{e.g.} \gls*{PINN}s) and mesh-based (\emph{e.g.} \gls*{FEM}) discretizations, making it agnostic to the specific method used to discretize the PDE. For example, we leverage the \gls*{FEM} formulation (that includes a customized mesh and preconditioners) to obtain the deformations in thin shells subject to gravity, providing estimates of the material properties and a framework for novel content generation. As we have shown, the use of our method with \gls*{PINN}s produces a very streamlined framework for learning the behavior of parametric PDEs.

While conventional \gls*{FEM} offers many advantages for solid mechanics problems, for other applications (such as fluid flows), other discretization techniques (such as immersed, particle-based methods~\cite{Ruberg:2011aa, lind2020review} or spectral methods) might be preferable. 
Furthermore, in current work, the variational inference is based on the mean field assumption, which is known to be overconfident in estimating the posterior uncertainty. The work can naturally be extended to utilize the connectivity structure in the \gls*{FEM} formulation to improve the calibration of the uncertainty estimates while leveraging the sparse structure imposed by the \gls*{FEM}~\cite{povala2022variational}. Other directions this work can be expanded is in the use of different architectures for the forward and inverse probabilistic maps, such as Fourier Neural Operators \cite{li2020fourier}, Wavelet Neural Operators \cite{tripura2022wavelet} or Physics-Informed Graph Neural Galerkin Networks \cite{gao2022physics}. This work could have significant impact in parametric PDE applications such as design and optimization scenarios and the study of complex systems.

%--------------------------------------------------------------------------------
\section*{Acknowledgements}
%--------------------------------------------------------------------------------
%
A.V. was supported by the Baxter \& Alma Ricard Foundation Scholarship. \"O.~D.~A. was partly supported by the Lloyd’s Register Foundation Data Centric Engineering Program and EPSRC Program Grant EP/R034710/1 (CoSInES). I.~K. was funded by a Biometrika Fellowship awarded by the Biometrika Trust. M.~G was supported by a Royal Academy of Engineering Research Chair, and EPSRC grants EP/R018413/2, EP/P020720/2, EP/R034710/1, EP/R004889/1, EP/T000414/1. F.~C. was supported by Wave 1 of The UKRI Strategic Priorities Fund under the EPSRC Grant EP/T001569/1, particularly the “Digital twins for complex engineering systems” theme within that grant, and The Alan Turing Institute.

\bibliographystyle{abbrv} 

\bibliography{biblio}

\newpage
\appendix
\addcontentsline{toc}{section}{Appendices}

%%%%%%%%%%%%%%%%%%%%%%%%%%%%%%%%%%%%%%%%%%%%%%%%%%%%%%%%%%%
% APPENDIX STARTS
%%%%%%%%%%%%%%%%%%%%%%%%%%%%%%%%%%%%%%%%%%%%%%%%%%%%%%%%%%%
\vbox{%
%  \hrule height 4pt
  \vskip 0.25in
  \vskip -\parskip%
    \hsize\textwidth
    % \linewidth\hsize
    \vskip 0.1in
    \centering
    {\LARGE\bf Appendix\par}
 \vskip 0.29in
  \vskip -\parskip
%   \hrule height 1pt
  \vskip 0.09in%
  }

%--------------------------------------------------------------------------------
\section{Experimental Details}
%--------------------------------------------------------------------------------
%
All experiments were run on an AMD Ryzen 9 5950X 16-Core Processor CPU and an NVIDIA GeForce RTX 3090 GPU. For all experiments, the TensorFlow GPU usage was limited between 6 and 13 GBs. The Poisson 1D examples are run on the CPU rather than the GPU as the \gls*{FEM} code is sequential and runs faster on the CPU. The 3D Bunny example was run on a mix of CPU and GPU. The PINN example was run entirely on the GPU as it can best leverage this type of hardware. We also note the definition used for the reported Mean Normalized Squared Error as 
\begin{equation}
    {\text{{MNSE}}}(x^{*}_{1:N}, x_{1:N}) := \frac{1}{N}\sum_i^N\frac{\|x^{*}_i - x_i\|^2}{\|x^{*}_i\|^2},
    \label{eq:MNSE}
\end{equation}
where $x^{*}$ denotes the base truth, and $x$ denotes the approximation.

%--------------------------------------------------------------------------------
\section{Alternative Derivation of {PDDLVM}}\label{sec:KLDivELBO}
%--------------------------------------------------------------------------------
%
In this section, we show how to derive the \gls*{PDDLVM} from the KL divergence between an intractable posterior and a variational approximation. We start by restating our joint model \eqref{eq:joint_probability_model_dataless} from Sect.~\ref{sec:prob_model:dataless} which includes the residual observation variable and the trainable inversion network
\begin{align}\label{eq:app:prob_model}
    p_\beta(\r,\ubf,\z, \fbf) = p(\r|\ubf,\z, \fbf)p_\beta(\z|\ubf, \fbf)p(\ubf)p(\fbf) .
\end{align}
We also restate our trainable variational approximation distribution ~\eqref{eq:joint_variational_app_dataless},
\begin{align}\label{eq:app:prob_var}
    &q_\alpha(\ubf, \z, \fbf) = q_\alpha(\ubf|\z,\fbf)p(\z)p(\fbf).
\end{align}
Using Bayes' rule we write for the intractable posterior of the latent variable given the observed residual
\begin{align}
    p_\beta(\ubf,\z,\fbf|\r) = \frac{p_\beta(\r,\ubf,\z,\fbf)}{p(\r)}.
\end{align}
We then consider the KL divergence between this intractable posterior and our tractable variational approximation
\begin{align}
    D_{KL}( q_\alpha(\ubf, \z, \fbf)||p_\beta(\ubf,\z,\fbf|\r)) &= \int \log \frac{q_\alpha(\ubf, \z, \fbf)}{p_\beta(\ubf,\z,\fbf|\r)}q_\alpha(\ubf, \z, \fbf)\md \ubf \md \z \md \fbf, \\
    &=\int \log \frac{q_\alpha(\ubf, \z, \fbf)p(\r)}{p_\beta(\r,\ubf,\z,\fbf)}q_\alpha(\ubf, \z, \fbf)\md \ubf \md \z \md \fbf \nonumber.
\end{align}
From this we factorize the evidence of the residual $p(\r)$ out of the integral
\begin{align}
    D_{KL}( q_\alpha(\ubf, \z, \fbf)||p_\beta(\ubf,\z,\fbf|\r)) = \int \log \frac{q_\alpha(\ubf, \z, \fbf)}{p_\beta(\r,\ubf,\z,\fbf)}q_\alpha(\ubf, \z, \fbf)\md \ubf \md \z \md \fbf +  \log p(\r) ,
\end{align}
while taking into account that $ \int q_\alpha(\ubf, \z, \fbf) \md \ubf \md \z \md \fbf \equiv 1 $. We then note that the KL divergence is a non-negative  quantity so that 
\begin{align}
     &D_{KL}( q_\alpha(\ubf, \z, \fbf)||p_\beta(\ubf,\z,\fbf|\r)) \geq 0.
\end{align}
This allows us to define the lower bound on the $\log$ marginal residual, as we further specify that $\r = \b0$. We then obtain a lower bound on the marginal likelihood of solving the parameterized PDE for all parameter values specified by their priors,
\begin{align}
    \log p(\r=\b0) \geq \int \log \frac{p_\beta(\r=\b0,\ubf,\z,\fbf)}{q_\alpha(\ubf, \z, \fbf)}q_\alpha(\ubf, \z, \fbf)\md \ubf \md \z \md \fbf.
\end{align}
Substituting the chosen factorizations \eqref{eq:app:prob_model} and \eqref{eq:app:prob_var} into the individual joint terms, we obtain the $\gls*{PDDLVM}$ framework derived in Sec.~\ref{sec:prob_model:dataless},
\begin{align}
    \log p(\r=\b0) \geq \int \log \frac{p(\r=\b0|\ubf,\z, \fbf)p_\beta(\z|\ubf, \fbf)p(\ubf)}{q_\alpha(\ubf|\z,\fbf)p(\z)}q_\alpha(\ubf|\z,\fbf)p(\z)p(\fbf)\md \ubf \md \z \md \fbf.
\end{align}
In order to relate the proposed probabilistic model to classical non-probabilistic approaches it is instructive to compare after introducing the probability densities and taking their logarithms the resulting expressions. 

%--------------------------------------------------------------------------------
\section{Additional Results for 1D Nonlinear Poisson}
\begin{figure}[h]
    \begin{subfigure}{\textwidth}
        \centering
        \includegraphics[width=1\textwidth]{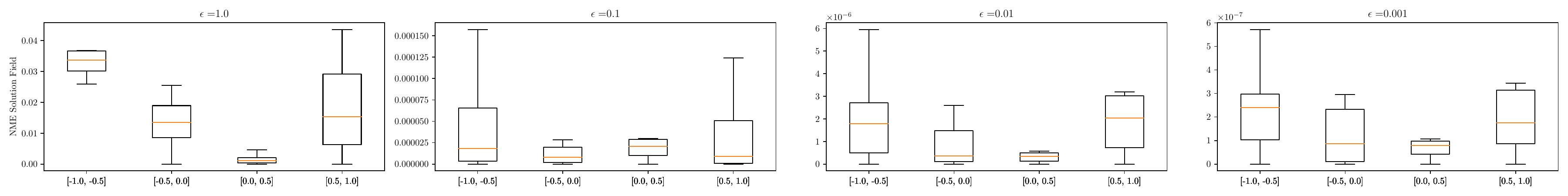}
        \caption{MNSE of $u(x)$.}
        \label{fig:poisson1DLinearRESIDUALU}
    \end{subfigure}
    \begin{subfigure}{\textwidth}
        \centering
        \includegraphics[width=1\textwidth]{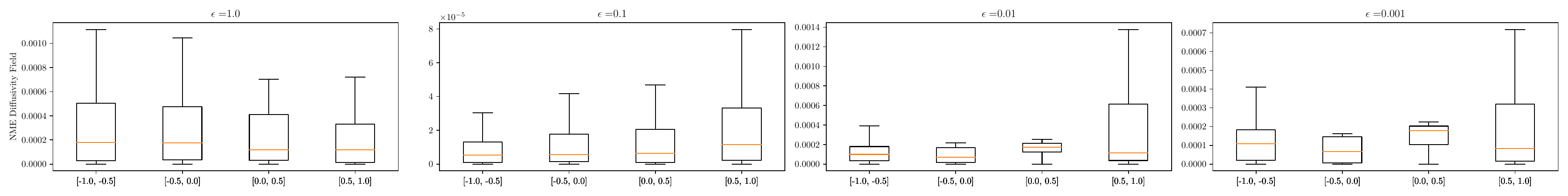}
        \caption{MNSE of the reconstructed $\kappa(x)$.}
        \label{fig:poisson1DLinearRESIDUALK}
    \end{subfigure}
    \begin{subfigure}{\textwidth}
        \centering
        \includegraphics[width=1\textwidth]{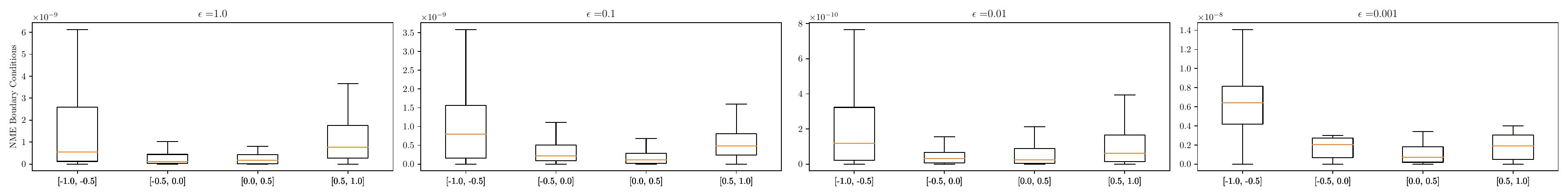}
        \caption{MNSE of $a$, and $b$ (boundary condition values).}
        \label{fig:poisson1DLinearRESIDUALB}
    \end{subfigure}
    \caption{The MNSE for different $\epsilon_{\r}$ values for the ranges of $\kappabf$ given on the x-axis. This was computed for the 1D linear Poisson example of Sec.~\ref{sec:LinearPoisson} and is computed from the same values as Fig.~\ref{fig:PoissonLinearLOG-MNSE} but not in the $\log$ scale. The prior on which the \gls*{FEM}-\gls*{PDDLVM} was trained was $p(\z) = \NPDF(0, 0.5)$.}
    \label{fig:PoissonLinearMNSEDistributionK}
\end{figure}

%--------------------------------------------------------------------------------
%
In Fig.~\ref{fig:PoissonLinearMNSEDistributionK}  we plot the same quantities as in Fig.~\ref{fig:PoissonLinearLOG-MNSE} but not in the log scale and in box plot format. We show the MNSEs for a variety of parameters of $\epsilon_{\r}$.
\begin{figure}[h!]
  \centering
  \includegraphics[width=0.8\linewidth]{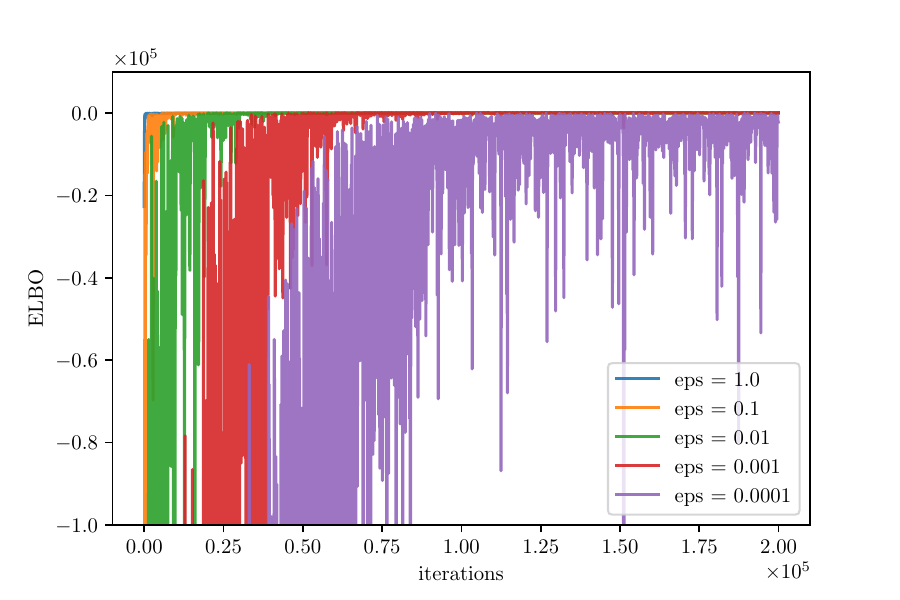}
  \caption{The convergence of the \gls*{ELBO} estimate for a 1D linear Poisson equation for various choices of $\epsilon_{\r}$.}
  \label{fig:Poisson1DELBOforEpsi}
\end{figure}
In Fig.~\ref{fig:Poisson1DELBOforEpsi} we plot the training of the \gls*{ELBO} for various settings of $\epsilon_{\r}$. We notice that the smaller the choice of $\epsilon_{\r}$, the slower the convergence, but the higher the accuracy.

%--------------------------------------------------------------------------------
\section{{PINN}-{PDDLVM}}\label{sec:AppendixPINNPDDLVM}
%--------------------------------------------------------------------------------
%
We stress the generality of using the weighted residual formulation, as noted by \cite{rixner2021probabilistic}, as different choices for the weight function $w(x)$ results in different well-known discretization methods. Setting
 $w_i(x) = \delta(x - x_i)$ we have a collocation method  \cite{strikwerda2004finite}, and for $w_i(x) = \boldsymbol{\delta}_{ij}  \text{ for }  x \in \Omega_j$ we obtain a subdomain (finite volume) method \cite{eymard2000finite}. For $w_i(x) = {\partial \cR(\hat{u}, \hdots)}/{\partial u_j}$ we have a least-squares method \cite{finlayson1966method}, for $w_i(x) = x^{m-1}$ we have a moment method (used in, \emph{e.g.}, electromagnetism \cite{gibson2021method}). For  $w_i(x) = \phi_i(x)$ we obtain a Galerkin method \cite{ern2004theory}. This could be a spectral method if $w_i(x) \neq 0 \text{ for every } x \in \Omega$ or \gls*{FEM} if $w_i(x) \neq 0 \text{ only for } x \in \Omega_k \subset \Omega$. 

In the case of \gls*{PINN} we clarify its link with the weighted residual method as a method of least-squares collocation \cite{finlayson2013method}. This is done by choosing $w_i(x) = \frac{\partial \cR(\hat{u}, x)}{\partial W}\delta(x_{i} - x)$ with $N_1,N_2,N_3$ collocation points in $\{\Omega, \Gamma, \mathcal{I}\}$, which are the domain, boundary conditions and initial conditions, respectively. $W$ denotes the collection of NN parameters. We abuse the notation and use $x =\{x, t\}$. As in \cite{finlayson1966method} we motivate the least-squares method as 
\begin{equation}
    r_i = \int_{\Omega}w_i(x)\cR_{\Omega}(\hat{u}, x)\md x + \int_\Gamma w_i(x)\cR_{\Gamma}(\hat{u}, x)\md x + \int_{\mathcal{I}} w_i(x)\cR_{\mathcal{I}}(\hat{u}, x)\md x.
\end{equation}
We then substitute the definition of $w_i(x) = \frac{\partial \cR(\hat{u}, x)}{\partial W}\delta(x_{i} - x)$,
\begin{align}
     r_i =  \int_{\Omega}\frac{\partial \cR_{\Omega}}{\partial W}\delta(x_{i} - x)\cR_{\Omega}(\hat{u}, x) & \md x +  \int_\Gamma \frac{\partial \cR_{\Gamma}}{\partial W}\delta(x_{i} - x) \cR_{\Gamma}(\hat{u}, x)\md x \nonumber\\
     & + \int_{\mathcal{I}} \frac{\partial \cR_{\mathcal{I}}}{\partial W} \delta(x_{i} - x) \cR_{\mathcal{I}}(\hat{u}, x)\md x.
\end{align}
After factorizing out the partial derivative with respect to $W$ and introducing the factor of one-half we obtain
\begin{align}
     r_i =  \frac{1}{2}\frac{\partial}{\partial W}\bigg[\int_{\Omega}\delta(x_{i} - x)\cR^2_{\Omega}(\hat{u}, x) & \md x +   \int_\Gamma \delta(x_{i} - x) \cR^2_{\Gamma}(\hat{u}, x)\md x \nonumber\\
     & +  \int_{\mathcal{I}}  \delta(x_{i} - x) \cR^2_{\mathcal{I}}(\hat{u}, x)\md x\bigg].
\end{align}
Since $r_i$ is required to be zero for all $i$, all residuals must be zero. We define
\begin{align}
     I_i(W) =  \cR^2_{\Omega}(\hat{u}, x_{i}) +  \cR^2_{\Gamma}(\hat{u}, x_{i}) +   \cR^2_{\mathcal{I}}(\hat{u}, x_{i}).
\end{align}
If we average over the points in each residual domain and assemble into a single objective, we obtain
\begin{equation}
    I(W) = \frac{1}{N_1}\sum_{x_{i}\in\Omega}\cR^2_{\Omega}(\hat{u}, x_{i}) +  \frac{1}{N_2}\sum_{x_i\in\Gamma} \cR^2_{\Gamma}(\hat{u}, x_{i}) + \frac{1}{N_3}\sum_{x_i\in\mathcal{I}} \cR^2_{\mathcal{I}}(\hat{u}, x_{i}),
\end{equation}
which is the commonly used training objective for \gls*{PINN}-type models.

\section{Additional Details}\label{app:info_tables}

In this subsection we collect information from the paper for easy referencing.

\textbf{1D Linear Poisson -- Sec.~\ref{sec:joint_learning_without_obs}}

\begin{table}[H]
\centering
\begin{tabular}{*{9}{c} }
    \toprule 
    dim $\ubf$ & dim $\r$   & dim $\z$ & $\z$ & $p(\z)$ & $\epsilon_\r$ & $\alpha$-Net & $\beta$-Net & iters\\
    5 & 21 & 1  & $\{\kappa\}$ & $\NPDF(0, 1)$ & $10^{-2}$  & $3l$, $50n$ & $3l$, $50n$ &  $2\times 10^{5}$\\   
    \bottomrule
\end{tabular}
\caption{Basic information 1D Linear Poisson}
\end{table}

\textbf{ 1D Linear Poisson Observable Map Inversion -- Sec.~\ref{sec:obs_map_inversion} }

\begin{table}[H]
\centering
\begin{tabular}{*{11}{c} }
    \toprule 
    dim $\ubf$ & dim $\r$   & dim $\z$ & dim $\fbf$ &  dim $\y$ & $\z$ & $\sigma_y$  & $\epsilon_\r$ & $\alpha$-Net & $\beta$-Net & iters\\
    8 & 101 & 4 & 1 & 80 & $\{\bkappa, a, b\}$ & $10^{-2}$ & $10^{-2}$ & $4l$, $50n$, & $4l$, $50n$, &  $1\times 10^{6}$\\   
    \bottomrule
\end{tabular}
\caption{Basic information 1D Linear Poisson Observable Map Inversion}
\end{table}

\begin{table}[H]
\centering
\begin{tabular}{*{4}{c} }
    \toprule 
    $p(\kappa_i)$ & $p(\fbf)$ & $p(a)$ & $p(b)$\\
    $\mathcal{U}(-1, 1)$ & $\mathcal{U}(1,5)$ & $\mathcal{U}(-0.5, 0.5)$ & $\mathcal{U}(-0.5, 0.5)$\\   
    \bottomrule
\end{tabular}
\caption{Prior information  1D Linear Poisson Observable Map Inversion}
\end{table}

% \newpage

\textbf{ 1D Nonlinear Poisson -- Sec.~\ref{sec:OneDPoissonNonlinear} }

\begin{table}[H]
\centering
\begin{tabular}{*{9}{c} }
    \toprule 
    dim $\ubf$ & dim $\r$   & dim $\z$  & dim $\fbf$ & $\z$ & $\epsilon_\r$ & $\alpha$-Net & $\beta$-Net & iters\\
    10 & 61 & 5  & 1 & $\{\bkappa, b\}$ & $10^{-2}$ & $4l$, $100n$ & $4l$, $100n$ &  $1\times 10^{6}$\\   
    \bottomrule
\end{tabular}
\caption{Basic information  1D Nonlinear Poisson}
\end{table}

\begin{table}[H]
\centering
\begin{tabular}{*{3}{c} }
    \toprule 
    $p(\z)$ & $p(\fbf)$ & $p(b)$ \\
    $\NPDF(0, 1)$ & $\mathcal{U}(1,1)$ & $\mathcal{U}(0.5, 1)$ \\
    \bottomrule
\end{tabular}
\caption{Prior information  1D Nonlinear Poisson}
\end{table}

\textbf{ Thin-Walled Flexible Shell -- Sec.~\ref{sec:3DshellObject} }

\begin{table}[H]
\centering
\begin{tabular}{*{9}{c} }
    \toprule 
    dim $\ubf$ & dim $\r$   & dim $\z$ & $\z$ & $p(\z)$ & $\epsilon_\r$ & $\alpha$-Net & $\beta$-Net & iters\\
    1029 & 24858 & 3  & $\{\bkappa\}$ & $\mathcal{U}(-2, 2)$ & $10^{-2}$  & $3l$, $2500n$ & $3l$, $2500n$ &  $5\times 10^{4}$\\   
    \bottomrule
\end{tabular}
\caption{Basic information Thin-Walled Flexible Shell}
\end{table}

\textbf{Nonlinear Heat Equation -- Sec.~\ref{sec:PINN_PDDLVM}}

\begin{table}[H]
\centering
\begin{tabular}{*{9}{c} }
    \toprule 
    dim $\ubf$ & dim $\r$   & dim $\z$ & $\z$ & $p(\z)$ & $\epsilon_\r$ & $\alpha$-Net & $\beta$-Net & iters\\
    $2\times 10^5$ & $2\times 10^5$ & 2 & $\{\gamma, \kappa\}$ & $\mathcal{U}(1, 5)\times \mathcal{U}(1, 5)$ & $10^{-2}$  & $4l$, $100n$ & $3c$, $1l-500n$ &  $5\times 10^{5}$\\   
    \bottomrule
\end{tabular}
\caption{Basic information Nonlinear Heat Equation}
\end{table}

\textbf{Inhomogeneous Wave Equation -- Sec.~\ref{sec:wave}}
\begin{table}[H]
\centering
\begin{tabular}{*{9}{c} }
    \toprule 
    dim $\ubf$ & dim $\r$   & dim $\z$ & $\z$ & $p(\z)$ & $\epsilon_\r$ & $\alpha$-Net & $\beta$-Net & iters\\
    $2.5\times 10^3$ & $2.5\times 10^3$ & 4 & $\{\fbf\}$ &  $\mathcal{U}(-5, 5)$ & $10^{-2}$  & $4l$, $100n$ & $3c$, $1l-500n$ &   $2.5\times 10^{5}$\\   
    \bottomrule
\end{tabular}
\caption{Basic information Inhomogeneous Wave Equation}
\end{table}

\end{document}